\documentclass{article}
\PassOptionsToPackage{numbers, compress}{natbib}
\usepackage[main,final]{neurips_2026}

\usepackage[utf8]{inputenc}
\usepackage[T1]{fontenc}
\usepackage{hyperref}
\usepackage{url}
\usepackage{booktabs}
\usepackage{multirow}
\usepackage{amsfonts}
\usepackage{nicefrac}
\usepackage{microtype}
\usepackage[table]{xcolor}
\usepackage{marvosym}
\usepackage{amsmath}
\usepackage{amssymb}
\usepackage{amsthm}

\usepackage{tabularx}
\usepackage{graphicx}
\usepackage{subcaption}
\usepackage{pifont}
\usepackage{enumitem}
\usepackage{algorithm}
\usepackage{algorithmic}
\usepackage{wrapfig}
\usepackage[most]{tcolorbox}
\usepackage[capitalize]{cleveref}
\crefname{section}{Sec.}{Secs.}
\Crefname{section}{Section}{Sections}
\Crefname{table}{Table}{Tables}
\crefname{table}{Tab.}{Tabs.}
\crefname{prop}{proposition}{propositions}
\Crefname{prop}{Proposition}{Propositions}

\newtheorem{prop}{Proposition}

\newcommand{\cmark}{\ding{51}}
\newcommand{\xmark}{\ding{55}}

\newcommand{\method}{LoHi}
\definecolor{takeaway_color}{HTML}{1F77B4}
\newtcolorbox{takeawaybox}[1]{%
    enhanced,
    breakable,
    colback=white,
    colframe=takeaway_color,
    boxrule=0pt,
    leftrule=5pt,
    arc=0pt,
    outer arc=0pt,
    sharp corners,
    left=8pt,
    right=2pt,
    top=3pt,
    bottom=3pt,
    fontupper=\small,
    notitle,
    before upper={\textcolor{takeaway_color}{\bfseries\small #1}\ },
}

\newcommand{\gmark}{\textcolor{black}{\cmark}}
\newcommand{\bmark}{\textcolor{gray!50}{\xmark}}
\definecolor{mygray}{gray}{0.85}
\definecolor{lohigreen}{rgb}{0.20, 0.55, 0.20}   
\definecolor{lohiyellow}{rgb}{0.85, 0.65, 0.13}  
\definecolor{lohired}{rgb}{0.70, 0.10, 0.10}     

\newcommand{\dotlow}{\textcolor{lohigreen}{$\bullet$}}
\newcommand{\dotmed}{\textcolor{lohiyellow}{$\bullet$}}
\newcommand{\dothigh}{\textcolor{lohired}{$\bullet$}}

\newcommand{\lesson}[1]{{\textcolor{takeaway_color}{L#1}}}

\hypersetup{
    colorlinks=true,
    citecolor=teal,
    linkcolor=red!70!black,
    urlcolor=magenta
}

\title{Rethinking Long-Video Efficiency: A Joint Allocation Perspective on Frames, Pixels, and Front-End Latency}

\author{%
  Sixun Dong\textsuperscript{1}
  Wei Li\textsuperscript{1}
  Andong Deng\textsuperscript{1}  
  Qi Qian\textsuperscript{2} 
  Victor Zhu\textsuperscript{3}  
  Zhengping Ji\textsuperscript{3} 
  Chen Chen\textsuperscript{1\Letter}\\[4pt]
  \textsuperscript{1}University of Central Florida  \qquad
  \textsuperscript{2}Meta Reality Labs  \qquad
  \textsuperscript{3}Axon  \\
  \texttt{\{sixundong,wei-li,chen.chen\}@ucf.edu}
}

\begin{document}

{\let\thefootnote\relax\footnotetext{\Letter~Corresponding author.}}

\maketitle

\begin{abstract}
Efficient long-video understanding with vision-language models (VLMs) has largely been treated as informative token selection at fixed native resolution: which frames or which visual tokens to retain under a token budget. We argue this framing leaves two axes unused~--- per-frame resolution itself can be traded for denser temporal coverage, and front-end decoding latency scales with the candidate pool, not the token budget. Through an empirical study across multiple VLMs and long-video benchmarks, we distill three lessons: (i) dense low-resolution sampling outperforms sparse native-resolution sampling at matched token budgets; (ii) some tasks are resolution-sensitive and benefit from high-resolution frames; and (iii) front-end decoding dominates wall time on hour-scale clips. These lessons motivate \method{}, a training-free, single-pass framework that pairs a dense low-resolution video stream with a sparse set of high-resolution image streams, processed through the VLM's native video and image pathways. 
Two lightweight selectors choose Hi-I frames: \method{}-Anchor uses codec I-frame metadata, and \method{}-SemDiv uses a query-relevance and visual-diversity DPP over CLIP features.
On three long-video benchmarks, \method{} improves over the vanilla native-resolution baseline by $+10.6$\% on average at matched token budget and over the strongest prior efficiency methods by $+5.2$\%, while reducing front-end decoding latency by up to $7\times$ on hour-scale clips. Project page: \url{https://sixundong.com/projects/lohi}.
\end{abstract}
\section{Introduction}
\label{sec:intro}

Recent vision-language models (VLMs) have made strong progress on static images and short videos~\cite{zhang2024llava,liu2024llavanext,zhang2025videollama,bai2025qwen3,wang2025internvl3_5}. Research is increasingly shifting toward long-video understanding, where task-relevant evidence is scattered across long temporal spans while decisive visual cues can be fleeting, resolution-sensitive, or dependent on long-term temporal relationships~\cite{tang2025adaptive,liu2025bolt,fan2026flashvid,song2024moviechat,shu2025video,dong2023weakly}.

Under a fixed visual-token budget, a long-video VLM can either sample more frames to achieve temporal coverage or use higher per-frame resolution to achieve spatial detail. However, existing efficient long-video VLMs largely treat this as an \emph{informative token selection} problem at fixed native resolution: which frames to select~\cite{radford2021learning,he2024ma,tang2025adaptive,zhang2025q,sun2025frames,zou2026air} or which visual tokens to retain~\cite{shen2025fastvid,dong2026mmtok,yang2025visionzip,fan2026flashvid,zhang2025sparsevlm} under the budget. This framing leaves two axes unused. First, per-frame resolution itself can be traded for temporal coverage~--- lowering resolution frees budget for more frames~--- yet resolution stays at native scale by default. Second, front-end latency, i.e., the time spent decoding the video and preprocessing frames before they reach the VLM, grows with video length and can exceed VLM inference itself, yet is rarely accounted for. 

The central question is therefore not which informative tokens to keep, but how to jointly allocate an end-to-end budget across frame count, per-frame resolution, and front-end latency. We revisit this sparse native-resolution recipe through an empirical study across multiple VLMs and long-video benchmarks (\cref{sec:lessons}), distilling three lessons.
\begin{figure*}[t]
  \centering
  \captionsetup[subfigure]{justification=centering,singlelinecheck=true}
  \begin{subfigure}[b]{0.31\linewidth}
    \centering
    \includegraphics[width=\linewidth]{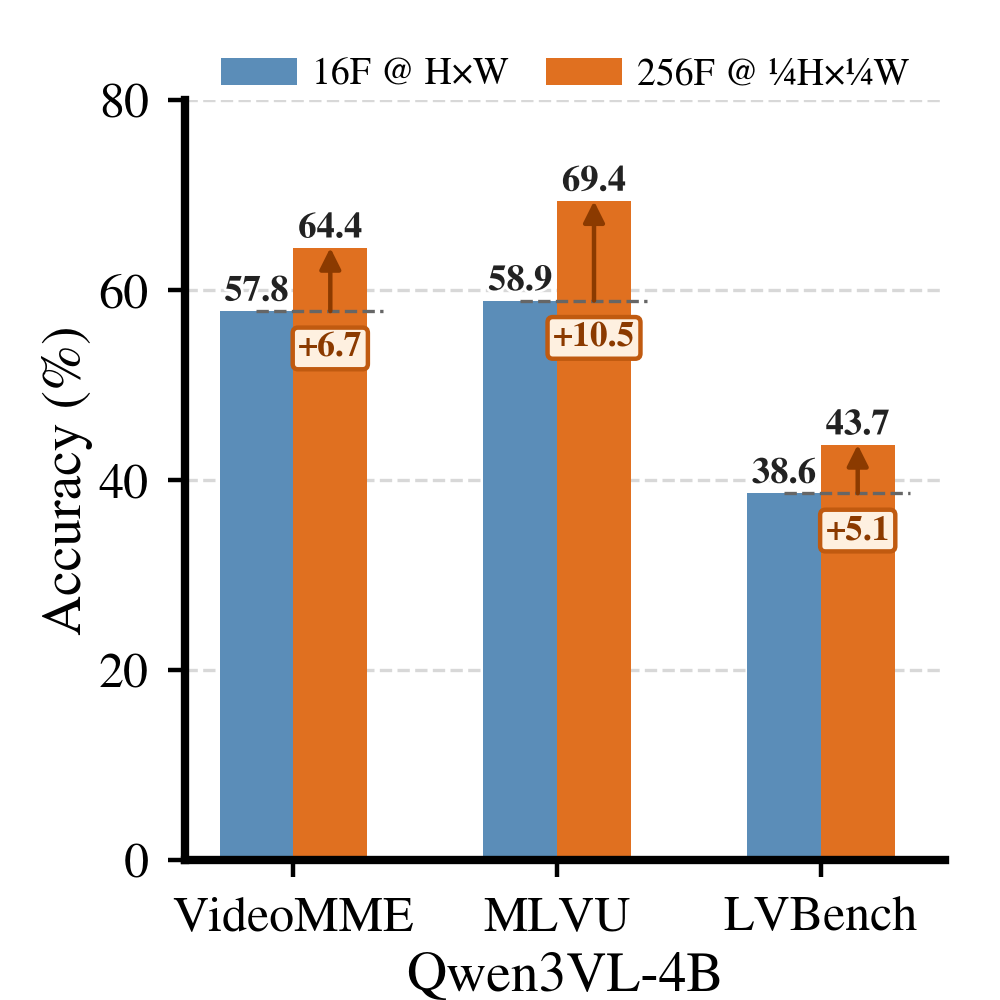}
    \caption{A new simple recipe}
    \label{fig:intro_o1}
  \end{subfigure}
  \hfill
  \begin{subfigure}[b]{0.31\linewidth}
    \centering
    \includegraphics[width=\linewidth]{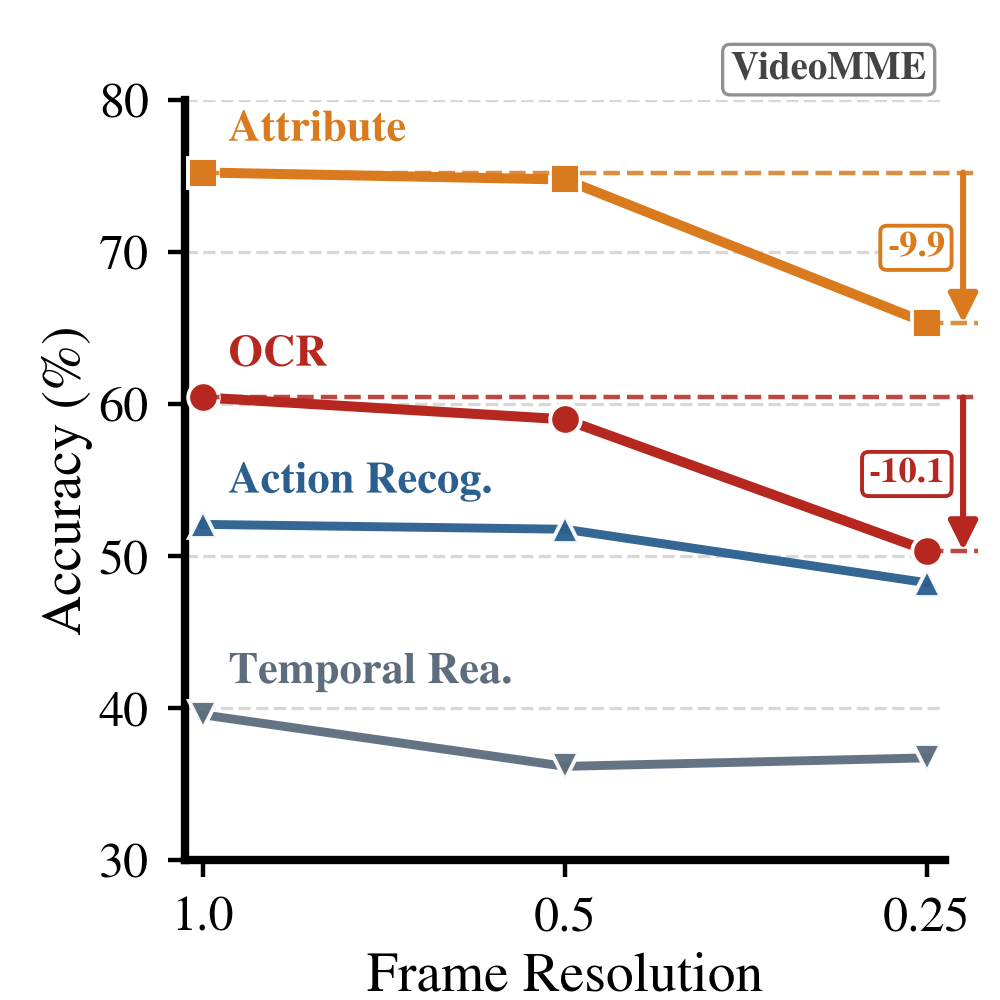}
    \caption{Resolution sensitivity}
    \label{fig:intro_o2}
  \end{subfigure}
  \hfill
  \begin{subfigure}[b]{0.31\linewidth}
    \centering
    \includegraphics[width=\linewidth]{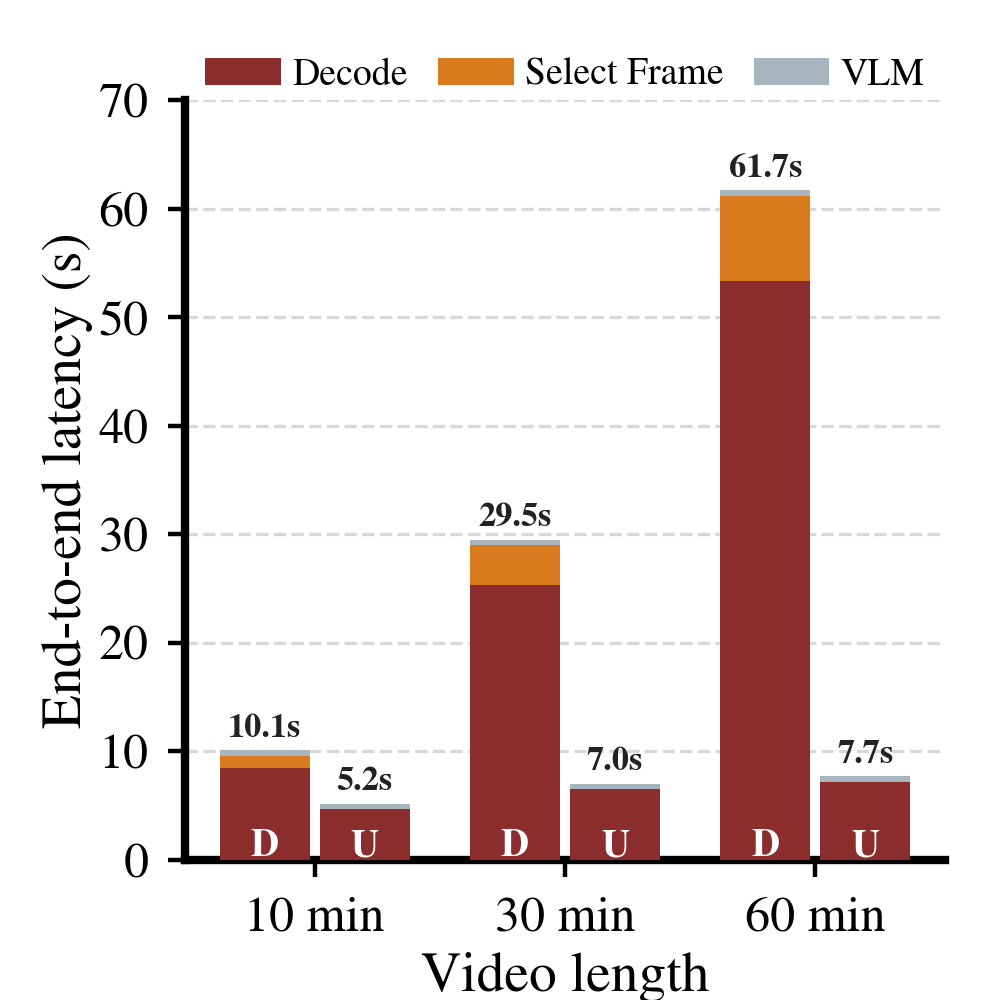}
    \caption{The overlooked decode latency}
    \label{fig:intro_o3}
  \end{subfigure}
    \caption{\textbf{Three practical lessons on long-video understanding.}
    \textbf{(a)} At matched token budgets, dense low-resolution sampling substantially outperforms the default baseline.
    \textbf{(b)} Resolution-sensitive tasks (OCR, Attribute Perception) significantly drop under aggressive down-scaling. Resolution-insensitive tasks (Action Recognition, Temporal Reasoning) remain robust.
    \textbf{(c)} Decode-then-select(D) scale linearly with video length; while uniform fixed-budget(U) keeps front-end latency constant.
    }
  \label{fig:intro_o123}
  \vspace{-18pt}
\end{figure*}

\begin{itemize}[leftmargin=1em, topsep=0.2em, itemsep=0.2em, parsep=0pt,partopsep=0pt]

  \item \textbf{L1. Dense low-resolution sampling is the new recipe.} At a fixed token budget, trading per-frame resolution for denser temporal coverage consistently outperforms the sparse native-resolution recipe on long-video benchmarks (\cref{fig:intro_o1}).

  \item \textbf{L2. Resolution sensitivity is task-dependent.} Moderate downscaling preserves most accuracy, but aggressive downscaling fails on resolution-sensitive tasks such as OCR and attribute perception (\cref{fig:intro_o2}). The remaining deficit on these tasks calls for a small amount of high-resolution evidence.
  
  \item  \textbf{L3. Front-end latency dominates on long videos.} Decode-then-select pipelines build their candidate pool by densely decoding the video, with latency scaling linearly with duration; on hour-scale videos this front-end latency dominates wall time, while uniform fixed-budget sampling stays constant (\cref{fig:intro_o3}).

\end{itemize}

Together, these lessons point to a simple paradigm for long-video understanding. Start from dense low-resolution temporal coverage, then add only a small amount of high-resolution image evidence needed to recover the remaining spatial gap, while keeping front-end latency low. We instantiate this idea in \textbf{LoHi} (\textbf{Lo}w-resolution video + \textbf{Hi}gh-resolution images), a training-free framework that separates temporal coverage from spatial detail. LoHi uses a dense \underline{\textbf{Lo}}w resolution \underline{\textbf{V}}ideo \textbf{(Lo-V)} base to cover the full timeline and adds only a sparse set of \underline{\textbf{Hi}}gh resolution \underline{\textbf{I}}mages \textbf{(Hi-I)} to recover fine spatial evidence. These two streams are processed through the VLM's native \emph{video} and \emph{image} pathways and linked by an inline text marker, requiring no fine-tuning or architectural changes. Beyond this initial instantiation, LoHi is highly extensible and exposes a modular design space over Hi-I selection, integration, and adaptive triggering. We make three contributions.

\begin{itemize}[leftmargin=1.5em, topsep=0.2em, itemsep=0.3em, parsep=0pt, partopsep=0pt]
  \item  We frame long-video VLM efficiency as a joint allocation problem over frame count, per-frame resolution, and front-end latency, supported by three empirical lessons across multiple VLMs and benchmarks.
 
  \item We propose \method{}, a training-free framework that addresses this allocation problem by composing a dense low-resolution video stream with a sparse high-resolution image stream through the VLM's native video and image pathways.

  \item \method{} improves accuracy over the strongest keyframe-selection and token-pruning baselines at matched token budget, while reducing front-end decoding latency by up to $7\times$ on hour-scale videos.
\end{itemize}
 
\section{Related Work}
\label{sec:related}
The frame--resolution allocation problem under strict token budgets remains the central challenge in long-video understanding. Existing approaches can be broadly categorized into fixed-resolution methods and resolution-aware methods. \Cref{tab:related-comparison} provides a qualitative comparison against \method{}.

\textbf{Fixed-resolution paradigms.}
To manage strict token budgets, existing efficient long-video approaches generally adopt one of two paradigms. The first is pre-encoder keyframe selection, which filters frames based on semantic relevance to the query (e.g., AKS~\cite{tang2025adaptive} adaptively partitions the timeline by query--frame similarity scores, and BOLT~\cite{liu2025bolt} uses inverse transform sampling). The second is post-encoder token pruning~\cite{yang2025visionzip,fan2026flashvid,zhang2025cdpruner}, which compresses visual features downstream of the vision encoder. By treating native per-frame resolution as a default, both paradigms severely limit the input frame count and sacrifice temporal coverage. Keyframe methods additionally incur front-end decode latency that scales with video length. \method{} breaks this assumption by decoupling temporal coverage (low-resolution video) from spatial detail (sparse high-resolution images).

\textbf{Resolution-aware exploration.}
A small set of recent works explores downscaling per-frame resolution. VisionThink~\cite{yang2025visionthink} lets the VLM decide whether high resolution is needed, adding autoregressive decoding overhead whose importance is highlighted by~\cite{dong2026rethinking}. MMTok~\cite{dong2026mmtok} pairs image-level downscaling with token pruning, but only on single images, leaving the temporal axis untouched. VideoLLaMA3~\cite{zhang2025videollama} introduces inter-frame compression, yet the temporal redundancy it exploits diminishes rapidly under sparse sampling. Q-frame~\cite{zhang2025q} dynamically adapts resolution for query-relevant frames, but still relies on dense semantic scoring per candidate frame. 
\method{} instead keeps the Lo-V stream at a fixed
resolution for temporal coverage and supplies spatial
detail through sparse Hi-I frames. Both streams are
processed in a single forward pass, without per-frame
resolution decisions within Lo-V.

\providecommand{\dotlow}{\textcolor{green!60!black}{\Large$\bullet$}}
\providecommand{\dotmed}{\textcolor{orange!80!black}{\Large$\bullet$}}
\providecommand{\dothigh}{\textcolor{red!70!black}{\Large$\bullet$}}
\providecommand{\gmark}{\textcolor{green!60!black}{\cmark}}
\providecommand{\bmark}{\textcolor{gray!65!black}{\xmark}}

\begin{table}[H]
\vspace{-10pt}
    \centering
    \small
    \caption{\textbf{Positioning LoHi in long-video efficiency.}
    Per-frame cost and Front-end latency are
    graded as low (\dotlow), medium (\dotmed), or high (\dothigh).
   }
    \label{tab:related-comparison}
    \small
    \setlength{\tabcolsep}{4pt}
    \resizebox{\linewidth}{!}{%
    \begin{tabular}{@{}l ccccc cc >{\columncolor{gray!10}}c@{}}
    \toprule
     & \multicolumn{5}{c}{\textit{Token-efficiency at fixed resolution}} & \multicolumn{2}{c}{\textit{Resolution-aware}} & Frames-Res\\
    \cmidrule(lr){2-6} \cmidrule(lr){7-8}
    \textbf{Axis}
      & \shortstack{FastV\\\cite{chen2024image}}
      & \shortstack{VisionZip\\\cite{yang2025visionzip}}
      & \shortstack{VideoLLaMA3\\\cite{zhang2025videollama}}
      & \shortstack{AKS\\\cite{tang2025adaptive}}
      & \shortstack{BOLT\\\cite{liu2025bolt}}
      & \shortstack{MMTok\\\cite{dong2026mmtok}}
      & \shortstack{VisionThink\\\cite{yang2025visionthink}}
      & \shortstack{\textbf{LoHi}\\\textbf{(ours)}} \\
    \midrule
    \textbf{Stage}
      & Post-encoder & Post-encoder & Post-encoder & Post-decode & Post-decode
      & Post-encoder & Multi-turn & \textbf{Pre-decode} \\
    \textbf{Selection signal}
      & Attn. & Sim. & Training. & Sim & Sim. & Coverage
    & LLM judgment & \textbf{Uniform} \\
    \textbf{Per-frame cost}
      & \dotmed & \dotmed & \dotmed & \dothigh & \dothigh
      & \dotmed & \dothigh & \dotlow \\
    \textbf{Front-end latency}
      & \dotlow & \dotlow & \dotlow & \dothigh & \dothigh
      & \dotlow & \dothigh & \dotlow \\
    \textbf{Modality}
      & Video & Video & Video & Video & Video & Image & Image & \textbf{V+I} \\
    \textbf{Resolution Tunes}
      & \bmark & \bmark & \bmark & \bmark & \bmark
      & \gmark & \gmark & \gmark \\
    \textbf{Train-free}
      & \gmark & \gmark & \gmark & \gmark & \gmark
      & \gmark & \bmark & \gmark \\
    \bottomrule
    \end{tabular}%
    }
\end{table}

\section{Lessons on Frames, Resolution, and Front-End Latency}
\label{sec:lessons}
Existing long-video VLMs allocate a fixed visual-token budget $B$ to a single, uniform sampling configuration $(N, r)$, where $N$ is the frame count and $r \in (0, 1]$ is the per-frame resolution scale relative to the native resolution $(r=1)$. Let $\tau(r)$ denote the number of visual tokens generated per frame at resolution scale $r$. Under the strict budget constraint $N \cdot \tau(r) \leq B$, models are forced into a resource-constrained trade-off between temporal density and spatial fidelity. We comprehensively evaluate this two-dimensional configuration space across multiple VLMs and benchmarks. Our empirical analysis yields three lessons that motivate a new paradigm for long-video understanding.

\subsection{Lesson 1 -- Dense, low-resolution sampling is the new recipe}
\label{sec:obs1}
\begin{figure*}[t]
  \centering
  \captionsetup[subfigure]{justification=centering,singlelinecheck=true}
  \begin{minipage}[c]{0.30\linewidth}
    \centering
    \includegraphics[width=\linewidth]{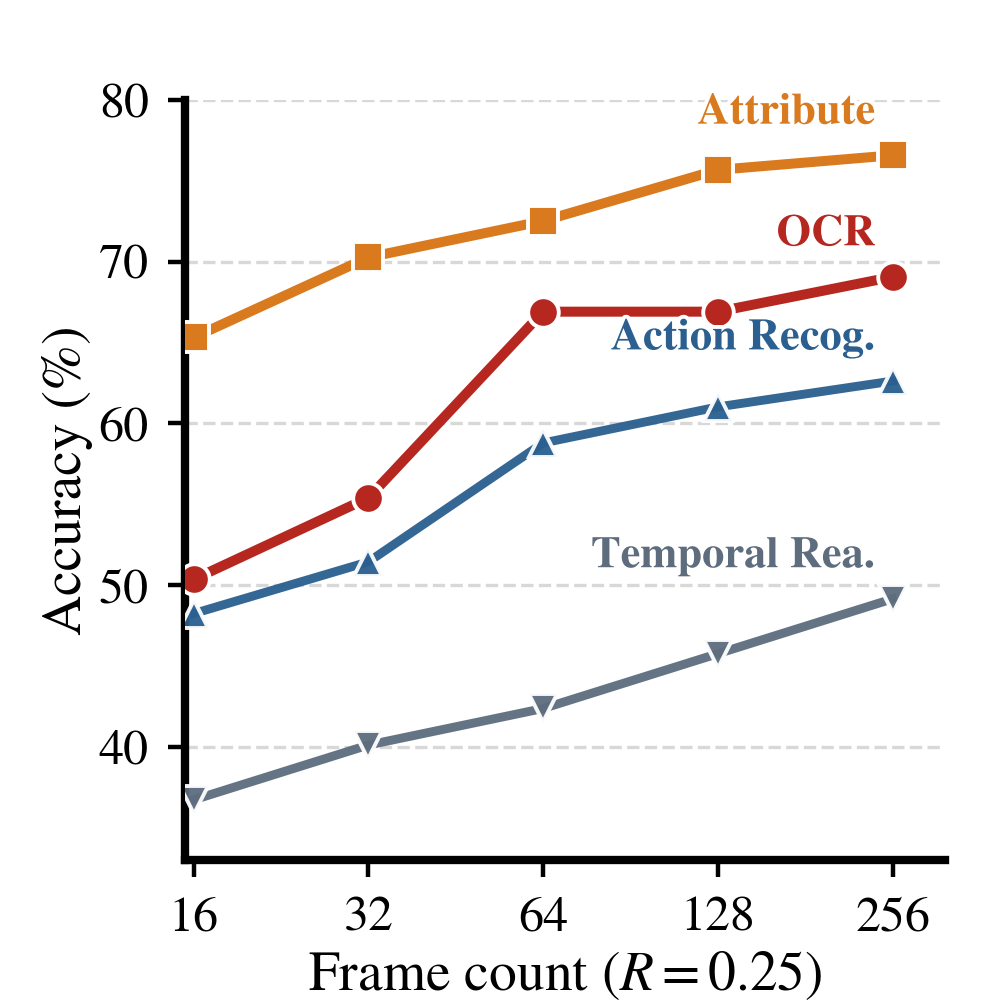}
    \subcaption{Frame density monotone gain.}
    \label{fig:obs2_pertask}
  \end{minipage}
  \begin{minipage}[c]{0.30\linewidth}
    \centering
    \includegraphics[width=\linewidth]{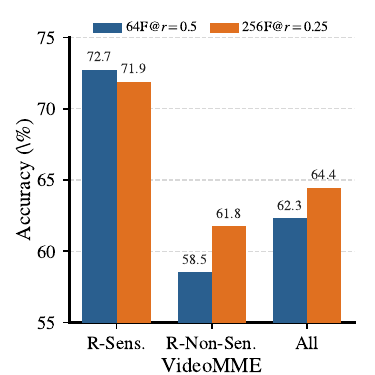}
    \subcaption{Frame--Resolution Allocation.}
    \label{fig:winner_flip}
  \end{minipage}
  \begin{minipage}[c]{0.38\linewidth}
    \centering
    \vspace{0pt}%
    \scriptsize
    \setlength{\tabcolsep}{1pt}
    \begin{tabular}{l l c c c}
      \toprule
      \shortstack[c]{Video\\Length}
        & \shortstack[c]{Frame\\Sample}
        & \shortstack[c]{GoP\\Seeks}
        & \shortstack[c]{Decoded\\Frames}
        & \shortstack[c]{Decode\\Latency} \\
      \midrule
      \multirow{2}{*}{10\,min} & Dense  & 600     & $\sim$18\,K   & 8.5\,s  \\
                               & Uniform  & 256     & $\sim$15\,K   & 4.7\,s  \\
      \midrule
      \multirow{2}{*}{30\,min} & Dense  & 1{,}800 & $\sim$54\,K   & 25.3\,s \\
                               & Uniform  & 256     & $\sim$27\,K   & 6.5\,s  \\
      \midrule
      \multirow{2}{*}{60\,min} & Dense & 3{,}600 & $\sim$108\,K  & 53.3\,s \\
                               & Uniform  & 256     & $\sim$27\,K   & 7.2\,s  \\
      \midrule
      \multirow{2}{*}{\tiny VideoMME}
                               & Dense & $\sim$272 & $\sim$24\,K & 14.6\,s \\
                               & Uniform  & $\sim$136 & $\sim$9\,K  & 3.9\,s  \\
      \bottomrule
    \end{tabular}
    \subcaption{Front-end decode cost.}
    \label{tab:seek_latency}
  \end{minipage}
  \small
    \caption{\textbf{Three lessons across frames, resolution, and front-end latency.} \textbf{(a)}~At fixed low resolution ($r{=}0.25$), accuracy on VideoMME rises monotonically with frame count $N$ grows from $16$ to $256$ for all four task categories (\cref{sec:obs1}).
    \textbf{(b)}~Under a matched token budget, resolution-sensitive tasks (R-Sens.) favor higher resolution over more frames, while resolution-insensitive tasks (R-Non-Sens.) prefer denser low-resolution sampling (\cref{sec:obs2}). \textbf{(c)}~Video decoder requires group-of-pictures (GoP) seeking, so constructing a dense candidate pool costs far more than uniform sampling(\cref{sec:lesson3}).
  }
  \label{fig:sec3_triptych}
\end{figure*}

\cref{tab:obs1_transfer} evaluates three frame--resolution allocations $(N, r)$ under the same fixed token budget (${\sim}5{,}760$ tokens): $(16, 1)$, $(64, 0.5)$, and $(256, 0.25)$. The advantage of dense, low-resolution sampling persists across settings: for Qwen3-VL-4B, $(256, 0.25)$ improves over the default baseline by $+6.66\%$ on VideoMME, $+10.47\%$ on MLVU, and $+5.10\%$ on LVBench, with the same trend on Qwen3-VL-8B, and on VideoLLaMA3-7B for MLVU and LVBench. 
\cref{fig:obs2_pertask} isolates the source of this gain: with resolution fixed at $r{=}0.25$, accuracy increases with $N$ from 16 to 256 for every VideoMME task category.
The gains are largest when the default recipe critically under-samples the timeline, and smaller when 16 frames already provide sufficient coverage. 
This establishes $(256, 0.25)$ as a strong global baseline; whether it generalizes across task types is the question of \cref{sec:obs2}.

\begin{table*}[ht]
\centering
\small
\caption{Comparison of diverse frontier VLMs on different long-video benchmarks.}
\label{tab:obs1_transfer}
\small
\resizebox{\linewidth}{!}{
\begin{tabular}{l ccc ccc ccc}
\toprule
\multirow{2}{*}{Setting} & \multicolumn{3}{c}{Qwen3-VL-4B} & \multicolumn{3}{c}{Qwen3-VL-8B} & \multicolumn{3}{c}{VideoLLaMA3-7B} \\
\cmidrule(lr){2-4} \cmidrule(lr){5-7} \cmidrule(lr){8-10}
& VideoMME & MLVU & LVBench & VideoMME & MLVU & LVBench & VideoMME & MLVU$^\dagger$ & LVBench \\
\midrule
{16}F${@}1.0$  & 57.78 & 58.89 & 38.61 & 60.04 & 58.95 & 38.22 & 46.81 & 35.88 & 32.09 \\
{64}F${@}0.5$    & 62.30 & 65.40 & 40.93 & 65.96 & 66.41 & 41.83 & \textbf{64.37} & 38.97 & 33.83 \\
{256}F${@}0.25$  & \textbf{64.44} & \textbf{69.36} & \textbf{43.71} & \textbf{67.04} & \textbf{71.29} & \textbf{44.80} & 50.04 & \textbf{40.30} & \textbf{36.99} \\
\bottomrule
\end{tabular}
}
{\footnotesize$^\dagger$ We report raw MLVU scores here;
results after answer-format normalization are provided
in \cref{sec:app_vl3_mlvu_norm}.}
\vspace{-10pt}
\end{table*}
\begin{takeawaybox}{Takeaway (L1)}
Under a strict token budget, trading per-frame resolution for dense temporal coverage consistently outperforms the default sparse, native-resolution recipe. Dense, low-resolution sampling should serve as the new foundational baseline for long-video VLMs.
\end{takeawaybox}
\vspace{-10pt}
\subsection{Lesson 2 -- Optimal frame-resolution allocation is task-dependent}
\label{sec:obs2}
Although \lesson{1}'s baseline performs well on average, the budget couples $N$ and $r$, forcing every operating point to trade temporal density against spatial fidelity. We refer to choosing along this trade-off as the \emph{frame--resolution allocation problem}, and ask whether a single point on it can serve all task types.
To answer this empirically, we rank VideoMME categories by their accuracy drop from $r{=}1.0$ to $r{=}0.25$ at $N{=}16$, and take the two categories with the largest drops (OCR and Attribute Perception) as the resolution-sensitive subset~(R), with the remaining categories as its complement~($\neg$R) (\cref{fig:intro_o2}). 
Under the same $\sim$5{,}760-token budget, the two operating points $(64, 0.5)$ and $(256, 0.25)$ swap dominance between the subsets (\cref{fig:sec3_triptych}\,(\textbf{b})): the higher-resolution configuration wins on R, while the higher-frame configuration wins on $\neg$R. The aggregate winner therefore reflects benchmark subset proportions rather than a uniform task-level optimum, and the L1 baseline, despite its average advantage, is suboptimal on R. No single $(N, r)$ pair can satisfy both subsets at the same budget.
\vspace{-5pt}
\begin{takeawaybox}{Takeaway (L2)}
Under a fixed token budget, no single $(N, r)$ optimally serves all task types; any globally fixed allocation is suboptimal on at least one subset.
\end{takeawaybox}

\subsection{Lesson 3 -- Front-end latency cannot be overlooked in long videos}
\label{sec:lesson3}

Modern video codecs~\cite{wiegand2003overview,punchihewa2020review} organize frames into Groups of Pictures (GoPs), where each GoP consists of a leading I-frame followed by a sequence of P/B-frames. An I-frame is a self-contained image and typically marks an abrupt motion or scene change, which is also where the codec inserts a new GoP boundary. For simplicity we focus on P-frames: a P-frame stores only the motion residual relative to its previous frame, ultimately chaining back to the I-frame. Sampling a P-frame inside a GoP therefore requires first decoding the I-frame and then sequentially decoding the preceding P-frames up to the target. For a policy that touches $G$ GoPs and samples $K$ frames at in-GoP depths $\{d_k\}_{k=1}^{K}$, the front-end decoding latency is approximately
{\small
\begin{equation}
    T_{\text{dec}} \;\approx\; G \cdot t_{I} \;+\; \Big(\sum_{k=1}^{K} d_k\Big) \cdot t_{P},
    \label{eq:decode-cost}
\end{equation}
}
where $t_I$ and $t_P$ denote the per-frame decoding latency of an I-frame and a P-frame, respectively. Decoding within each GoP is strictly sequential; including B-frames, which additionally reference future frames, only worsens this picture.

\textbf{The overlooked decode latency.} We profile Eq.~\eqref{eq:decode-cost} on videos of three lengths (10/30/60\,min) and on VideoMME (\cref{fig:sec3_triptych}\,(\textbf{c})). Dense sampling  (FPS$=$1) drives both $G$ and $\sum_k d_k$ linearly with duration, whereas uniform sampling at $N{=}256$ caps both. As a result, the latency gap widens monotonically with video length, reaching $7.4\times$ on 60-min videos, and remains substantial on VideoMME itself, confirming this is not a stress-test artifact. This is precisely the cost paid by decode-then-select methods such as keyframe selection~\cite{tang2025adaptive,liu2025bolt}, which build their candidate pool by densely decoding the video at FPS$=$1 in advance~--- a substantial front-end latency consistently overlooked by prior work. Method design and latency evaluation should therefore account for this front-end latency rather than report only the model-side token budget.
\vspace{-5pt}
\begin{takeawaybox}{Takeaway (L3)}
Under dense candidate-pool decoding such as FPS$=$1, front-end latency grows with video length and eventually dominates end-to-end runtime. Long-video efficiency methods must therefore bound the size of the decoded candidate pool, not just the visual-token budget passed to the VLM.
\end{takeawaybox}
\vspace{-10pt}
\section{The LoHi Framework}
\label{sec:method}

We unify the lessons from \cref{sec:lessons} into a single problem statement: \emph{frame--resolution allocation under a front-end latency constraint}. We first formalize it via a cross-resolution decomposition (\cref{sec:method-decomp}), then instantiate it as the training-free single-pass framework \textbf{LoHi} (\cref{sec:method-lohi}), enhance its high-resolution stream with a diversity-aware selector (\cref{sec:semdiv}), and finally add an entropy-triggered adaptive variant (\cref{sec:adaptive}).

\subsection{Cross-Resolution Decomposition}
\label{sec:method-decomp}
To address the frame--resolution allocation problem (\cref{sec:obs2}), we propose a \emph{cross-resolution decomposition}: split the single configuration $(N, r)$ into two complementary streams sharing one budget. A dense \emph{low-resolution stream} of $N$ frames at scale $r_\ell$ provides temporal coverage (\lesson{1}); a sparse \emph{high-resolution stream} of $K \ll N$ frames at scale $r_h > r_\ell$ supplies spatial detail (\lesson{2}). The $K$ Hi-I indices are drawn as a subset of the Lo-V grid $[N]$, so each frame is decoded once at native resolution and reused at both scales. The two streams jointly satisfy a token budget and a decode budget:
\begin{equation}
\underbrace{N \cdot \tau(r_\ell) \;+\; K \cdot \tau(r_h) \;\leq\; B}_{\text{token budget (\lesson{1}, \lesson{2})}}, 
\qquad 
\underbrace{N_{\text{dec}} \;=\; N}_{\text{decode budget (\lesson{3})}}.
\label{eq:dual-stream}
\end{equation}
VLM compute stays bounded by $B$, identical to a single-stream baseline. Front-end decode count $N_{\text{dec}}$ depends only on the chosen $N$, not on video length: each of the $N$ frames is decoded once at native resolution, then resized to $r_\ell$ for Lo-V and additionally kept at $r_h$ for the $K$ selected Hi-I. Compared with a single-stream baseline at $(N_{\text{base}}, r_{\text{base}})$ that decodes $N_{\text{base}} > N$ frames, e.g.,(256F,0.25), our decomposition saves $(N_{\text{base}} - N)$ decoded frames at the same token budget.

\begin{figure}[t]
\centering
\IfFileExists{fig/lohi_framework.jpg}{%
  \includegraphics[width=\linewidth]{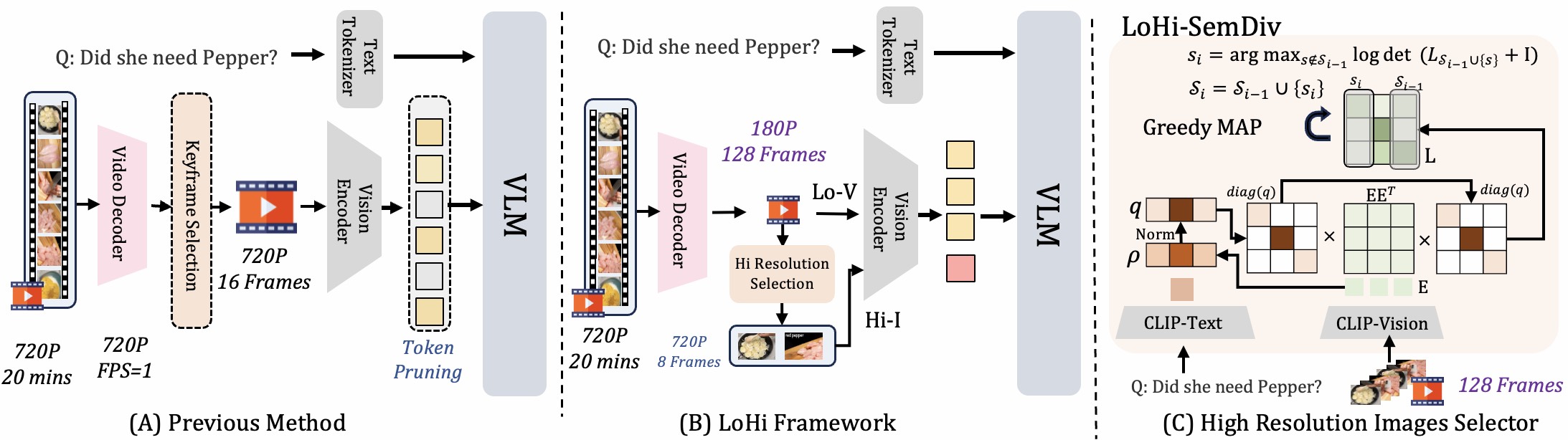}%
}{%
  \fbox{\parbox{0.9\linewidth}{\centering [Figure~\ref{fig:architecture} placeholder] LoHi pipeline figure.}}%
}
\caption{\textbf{Overview of LoHi framework.} Compared with previous work (a) which prunes or selects within a native resolution pathway, LoHi (b) decomposes one decoded video into two complementary streams: dense low-resolution \emph{Lo-V} via the video pathway and sparse high-resolution \emph{Hi-I} via the image pathway. (c) shows the LoHi-SemDiv selector, which picks Hi-I indices via greedy MAP.}
\label{fig:architecture}
\vspace{-10pt}
\end{figure}

\subsection{LoHi: Instantiating the Decomposition via Video and Image Pathways}
\label{sec:method-lohi}

Modern unified video-image VLMs (e.g., Qwen-VL series~\citep{bai2025qwen3}) expose two input pathways sharing the same vision tower but differing in positional encoding: a \emph{video pathway} with 3D temporal-spatial mRoPE \cite{su2024roformer,wang2024qwen2}, and an \emph{image pathway} treating each input as an independent 2D image. We instantiate the decomposition of Sec.~\ref{sec:method-decomp} by routing the low-resolution stream through the video pathway and the high-resolution stream through the image pathway. mRoPE on the video pathway assigns each frame a globally consistent temporal position so motion is preserved at its natural rate; the image pathway gives each high-resolution image its own 2D positional encoding, leaving the video timeline undisturbed. 
This yields \textbf{LoHi}, a two-stream framework that combines a low-resolution video stream (\textbf{Lo-V}) with a high-resolution image stream (\textbf{Hi-I}).

We then bridge these two streams via a natural-language marker to tell the VLM the relationship between a high-resolution image and a low-resolution video. 
\vspace{-3pt}
\begin{quote}\small\itshape
[Video] The above low-resolution video provides temporal context. The following $K$ high-resolution images show selected key frames: [img$_1$], \ldots, [img$_K$].
\end{quote}
\vspace{-5pt}

\textbf{LoHi-Uniform.} Serving as the fundamental baseline of our method, LoHi-Uniform simply selects the $K$ Hi-I frames at regular intervals directly from the Lo-V grid. By relying solely on mathematical indexing, this strategy incurs strictly zero additional computational overhead.

\textbf{Codec-guided Hi-I selector.} As established in \cref{sec:lesson3}, video encoders naturally anchor GoPs with I-frames, which can align with abrupt scene changes or significant motion. We propose a codec-aware selector, \textbf{LoHi-Anchor}, which explicitly exploits these pre-computed boundaries as a free semantic signal. 
It reads pre-computed I-frame indices without additional pixel decoding. For each uniformly spaced position $u_m$, it finds the nearest I-frame and selects the Lo-V grid frame closest to it. All Hi-I frames therefore come from the same decoded candidate pool (see \cref{sec:app_anchor} for details). 

While LoHi-Uniform and LoHi-Anchor both achieve negligible-cost selection by avoiding pixel-level scoring, they do not explicitly guarantee the visual distinctiveness of the sampled frames. To address this, we introduce a third selector that explicitly optimizes for semantic diversity.

\subsection{LoHi-SemDiv: Diversity-Aware Selection via Quality-Similarity DPP}
\label{sec:semdiv}

Inspired by the subset-selection framing of~\citep{dong2026mmtok}, we cast Hi-I selection as choosing $K$ frames from the Lo-V grid $[N]$ given a query $Q$. Unlike MMTok, whose objective is to \emph{cover} the original token sets, temporal coverage is already inherently supplied by our Lo-V base. Therefore, the $K$ sparse Hi-I slots must instead be (i) \emph{mutually diverse}, contributing distinct spatial evidence, and (ii) \emph{query-relevant}, allocating the strict budget exclusively to content informative for $Q$.

The quality--similarity decomposition of determinantal point processes (DPPs)~\cite{kulesza2012determinantal} elegantly unites these two criteria within a single submodular objective: it promotes relevance via the quality term while penalizing redundancy via the similarity kernel. While recent work~\cite{zhang2025cdpruner} applies DPPs to token-level pruning within a single static image, we elevate this formulation to the temporal dimension, operating at frame granularity across two complementary resolution streams. We instantiate this diversity-aware, cross-stream selector on the Lo-V grid as \textbf{LoHi-SemDiv}. A detailed complexity analysis is provided in ~\cref{sec:app_semdiv_alg}.

\textbf{DPP kernel.} Let $e_j \in \mathbb{R}^d$ denote the L2-normalized CLIP \cite{radford2021learning} embedding of frame $j$, and let $\rho_j = \cos\bigl(\mathrm{CLIP}_\mathrm{txt}(Q),\, e_j\bigr)$ be the raw query--frame cosine similarity. We min-max normalize and power-sharpen $\rho_j$ into a per-frame quality score $q_j$, then couple per-frame relevance with pairwise visual similarity via the quality--similarity decomposition\cite{kulesza2012determinantal}:
\begin{equation}
\tilde{\rho}_j = \frac{\rho_j - \min_m \rho_m}{\max_m \rho_m - \min_m \rho_m},
\qquad
q_j = \tilde{\rho}_j^{\,\alpha},
\qquad
L_{ij} = q_i\, q_j\, e_i^\top e_j.
\label{eq:semdiv_kernel}
\end{equation}
Diagonals $L_{jj} = q_j^2$ encode per-frame relevance, off-diagonals couple relevance with pairwise visual similarity, so subsets of similar frames yield small $\det(L_\mathcal{S})$. Stacking the embeddings as $E = [e_1, \ldots, e_N]^\top \in \mathbb{R}^{N \times d}$ and qualities as $\mathbf{q} = (q_1, \ldots, q_N)^\top$, the kernel admits the equivalent matrix form
\begin{equation}
L = \mathrm{diag}(\mathbf{q})\, E E^\top\, \mathrm{diag}(\mathbf{q}),
\label{eq:semdiv_kernel_matrix}
\end{equation}
which matches the construction in \cref{fig:architecture}\,(c) and is positive semi-definite. We  set $\alpha = 5$.

\textbf{Selection objective.} For a theoretically grounded greedy approximation, we adopt the regularized log-determinant:
\begin{equation}
\mathcal{S}^* = \arg\max_{|\mathcal{S}| = K} g(\mathcal{S}),
\qquad
g(\mathcal{S}) = \log\det\bigl(L_\mathcal{S} + \mathbf{I}\bigr).
\label{eq:semdiv_obj}
\end{equation}
The identity regularization ensures all eigenvalues of $L_\mathcal{S} + \mathbf{I}$ are at least one, so $g$ is non-negative, monotone non-decreasing, and submodular~\cite{ kulesza2012determinantal}. This admits a near-optimal greedy approximation:
\begin{prop}[\citep{nemhauser1978analysis}]
\label{prop:semdiv_greedy}
Let $\mathcal{S}$ be the subset obtained by greedy maximization of $g$. Then
\begin{equation*}
g(\mathcal{S}) \;\geq\; (1 - 1/e)\,\max_{|\mathcal{A}|=K} g(\mathcal{A}).
\end{equation*}
\end{prop}

\vspace{-5pt}


\subsection{Adaptive LoHi via Entropy Triggering} 
\label{sec:adaptive}
While the Hi-I stream provides crucial spatial details, processing $K$ high-resolution frames incurs additional computational overhead. However, because many queries can be answered correctly using the Lo-V base alone, our decoupled two-stream architecture naturally supports an adaptive, per-question early-exit mechanism. Specifically, we first evaluate the query using the lightweight Lo-V base and compute the entropy of the generated answer distribution. The Hi-I stream is triggered only if the model's uncertainty exceeds a pre-defined threshold $\theta$:
\begin{equation}
\text{Trigger Hi-I} \iff H\big(p_{\text{Lo-V}}(\cdot \mid Q)\big) > \theta.
\label{eq:multiturn}
\end{equation}
This entropy-triggered protocol reduces the average computational cost by bypassing the Hi-I stream for simple queries. As detailed in \cref{sec:app_multiturn}, this adaptive variant nearly matches the performance of the full \method{} pipeline while invoking the high-resolution stream on only one-third of the queries.
\section{Experiments}
\label{sec:experiments}
\paragraph{Setup.} We evaluate \method{} on three long-video benchmarks---VideoMME (without subtitles)~\citep{fu2025videomme}, MLVU~\citep{zhou2025mlvu}, and LVBench~\citep{wang2025lvbench}. Additional results with subtitles are provided in \cref{sec:app_subtitles}. We employ Qwen3-VL-4B~\citep{bai2025qwen3} as the primary backbone and evaluate generalization on
Qwen3-VL-8B (\cref{sec:app_lohi_qwen8b});
Qwen3.5-4B~\citep{qwen3.5} (\cref{sec:app_qwen35});
and VideoLLaMA3-7B~\citep{zhang2025videollama},
VideoChat3-4B~\citep{li2026videochat3}, and
Qwen2.5-VL-7B~\citep{bai2025qwen2}
(\cref{sec:more_vlms}). We compare against representative keyframe-selection methods (AKS~\citep{tang2025adaptive}, BOLT~\citep{liu2025bolt}) and two token-pruning methods (FlashVid~\citep{fan2026flashvid} for video VLMs and VisionZip~\citep{yang2025visionzip} for image VLMs). We also include CLIP-Topk as a keyframe-selection baseline. 

\paragraph{Implementation details.}
Following \lesson{3}, all methods cap the decoded frames per video at $N_{\text{dec}} \!\leq\! 256$ and share the visual-token budget of the default $(16, 1.0)$ recipe ($\sim$5{,}760 tokens). Key-frame selection baselines decode $256$ frames at $r{=}1.0$ and pick $16$; token-pruning baselines decode $256$ at $r{=}1.0$ and prune $93.75\%$ of the tokens. \method{} uses $N{=}128$ at $r_\ell{=}0.25$ for Lo-V and $K{=}8$ Hi-I indices drawn from these $128$ at $r_h{=}1.0$, giving $N_{\text{dec}}{=}128$~--- half of the baselines at the same token budget. We denote this as $128$F@$0.25$+Hi-I. Full details are in ~\cref{sec:app_implementation}.

\subsection{Performance Comparison across different benchmarks}
\label{sec:main_result}

\paragraph{Comparison on Qwen3-VL-4B.} \cref{tab:main} reports accuracy at a default token budget. \method{}-SemDiv achieves the best performance on all three benchmarks, reaching $66.81$\,/\,$74.38$\,/\,$45.90\%$ on VideoMME\,/\,MLVU\,/\,LVBench, an average gain of $+10.60\%$ over the vanilla baseline (16F,1.0), and $+5.23\%$ over the strongest prior baseline (BOLT\cite{liu2025bolt}) at identical token cost. Notably, \textbf{Low-Res-Base}, the dense low resolution video stream ($256$F$@0.25$) already outperforms every prior keyframe-selection and token-pruning baseline on average, supporting \lesson{1} that varying $r$ alone provides a competitive recipe at fixed budget without any selection logic. Adding Hi-I extends the lead, with \method{}-SemDiv $>$ \method{}-Anchor $>$ \method{}-Uniform on every benchmark and the largest semantic-diversity contribution on MLVU ($+6.76\%$ over \method{}-Uniform). Paired significance tests and task-level analyses are
provided in \cref{sec:app_selection_analysis}. Crucially, \method{} decodes only $128$ frames per video---half the $256$-frame pool consumed by every prior efficiency baseline. Specifically, \method{}-Uniform and \method{}-Anchor rely entirely on simple frame indices or free metadata, which is almost zero cost. \method{}-SemDiv adds one CLIP forward pass over the $128$ low-resolution frames and has lower end-to-end latency than token pruning and keyframe-selection methods (see \cref{tab:ttft_videomme} for details). We also evaluate our method on Qwen3-VL-8B in ~\cref{sec:app_lohi_qwen8b}.

\begin{table}[t]
    \centering
    \small
    \caption{Performance Comparison on Qwen3-VL-4B. Avg.\ is the mean over three benchmarks.}
    \label{tab:main}
\resizebox{\linewidth}{!}{
    \begin{tabular}{l c c c c c c c c c c}
    \toprule
    \multirow{2}{*}{Method} & \#Frames & Setting & Tokens & \multicolumn{4}{c}{VideoMME (\%)} & MLVU & LVBench & Avg. \\
    \cmidrule(lr){5-8}
     & (Decoded) & ($N$F${@}r$) & (Budget) & Short & Med. & Long & All & (\%) & (\%) & (\%) \\
    \midrule
    Qwen3-VL-4B  & 16 & $16$F${@}1.0$ & 5{,}760 & 68.44 & 55.67 & 49.22 & 57.78 & 58.89 & 38.61 & 51.76 \\
    \midrule
    \rowcolor{mygray}
    \multicolumn{11}{l}{\emph{The new recipe: dense low-resolution video}} \\
    Low-Res-Base & 256 & $256$F${@}0.25$ & 5{,}760 & 75.78 & 62.33 & 55.22 & 64.44 & 69.36 & 43.71 & 59.17 \\
    \midrule
    \rowcolor{mygray}
    \multicolumn{11}{l}{\emph{Pre-encoder keyframe selection}} \\
    CLIP-TopK & 256 $\rightarrow$ 16 & $16$F${@}1.0$ & 5{,}760 & 65.67 & 56.33 & 49.56 & 57.19 & 69.36 & 44.74 & 57.10 \\
    AKS \cite{tang2025adaptive} & 256 $\rightarrow$ 16 & $16$F${@}1.0$ & 5{,}760 & 65.89 & 56.22 & 49.67 & 57.26 & 69.22 & 44.67 & 57.05 \\
    BOLT \cite{liu2025bolt} & 256 $\rightarrow$ 16 & $16$F${@}1.0$ & 5{,}760 & 71.44 & 59.00 & 52.00 & 60.81 & 68.11 & 42.48 & 57.13 \\
    \midrule
    \rowcolor{mygray}
    \multicolumn{11}{l}{\emph{Post-encoder token pruning. Pruning Ratio:93.75\% to keep same token budget}} \\
    VisionZip\cite{yang2025visionzip} & 256 & $256$F${@}1.0$ & 5{,}760 & 64.78 & 55.67 & 50.56 & 57.00 & 62.46 & 35.13 & 51.53 \\
    FlashVid\cite{fan2026flashvid} & 256 & $256$F${@}1.0$ & 5{,}760 & 69.56 & 58.11 & 50.22 & 59.30 & 63.70 & 34.09 & 52.36 \\
    \midrule
    \rowcolor{mygray}
    \multicolumn{11}{l}{\emph{LoHi (ours)}} \\
    LoHi-Uniform & 128 & $128$F${@}0.25+$Hi-I & 5{,}760 & 75.67 & \textbf{66.44} & 55.22 & 65.78 & 67.62 & 44.09 & 59.16 \\
    LoHi-Anchor  & 128 & $128$F${@}0.25+$Hi-I & 5{,}760 & 76.00 & 66.22 & 55.44 & 65.89 & 68.40 & 44.42 & 59.57 \\
    LoHi-SemDiv  & 128 & $128$F${@}0.25+$Hi-I & 5{,}760 & \textbf{77.11} & 65.78 & \textbf{57.56} & \textbf{66.81} & \textbf{74.38} & \textbf{45.90} & \textbf{62.36} \\
    \bottomrule
    \end{tabular}
    }
\end{table}
\begin{figure}[t]  
 
    \vspace{-5pt}
    \begin{minipage}{0.43\linewidth}
      \centering
      \captionof{table}{ Frame-Resolution Allocation.}
      \label{tab:rsplit}
      \small
      \setlength{\tabcolsep}{2pt}
      \resizebox{\linewidth}{!}{
      \begin{tabular}{l c c c c}
      \toprule
      Qwen3-VL-4B & Tokens & R (\%) & $\neg$R (\%) & All (\%) \\
      \midrule
      \rowcolor{gray!20}
      \multicolumn{5}{l}{\emph{Under Single Resolution and Frames setting $(N,r)$.}} \\
      64F@0.5 & 5{,}760 & \underline{72.73} & 58.54 & 62.30 \\
      256F@0.25 & 5{,}760 & 71.89 & \underline{61.76} & \underline{64.44} \\
      \midrule
      \rowcolor{gray!20}
      \multicolumn{5}{l}{\emph{LoHi with Hi-I Selector}} \\
      LoHi-Uniform & 5{,}760 & 76.64 & 61.86 & 65.78 \\
      LoHi-Anchor & 5{,}760 & 75.38 & 62.47 & 65.89 \\
      LoHi-SemDiv & 5{,}760 & \textbf{77.06} & \textbf{63.12} & \textbf{66.81} \\
      \bottomrule
      \end{tabular}
      }
    \end{minipage}
    \hfill  
  \begin{minipage}{0.55\linewidth}
      \centering
       \captionof{table}{Comparison of Inference Efficiency.}
      \label{tab:ttft_videomme}
      \small
      \setlength{\tabcolsep}{2pt}  
      \resizebox{\linewidth}{!}{
      \begin{tabular}{l | c | c | c | c | c | c | c}
      \toprule
      \multirow{2}{*}{Method} & {Decode} & {Sel.} & {ViT} & {TTFT} & \multirow{2}{*}{TFLOPs} & {Mem.} & {VideoMME} \\
       & s & ms & ms & ms &  &  +8.3GB & ACC \\
      \midrule
      256F@0.25    & 5.4 & 50    & 216   & 5,963 & 82.2  & 1.3 & 64.44 \\
      Key-frame    & 5.4 & 130   & 151   & 5,978 & 89.3  & 1.3 & 60.81 \\
      VisionZip    & 5.4 & 4     & 2,486 & 8,186 & 464.3 & 15.4 & 57.00 \\
      FlashVid     & 5.4 & 808   & 2,486 & 8,990 & 464.3 & 15.4 & 59.30 \\
      \midrule
      LoHi-Uniform & 3.3 & 51    & 195   & \textbf{3,836} & 84.6  & 1.3 & 65.78 \\
      LoHi-Anchor  & 3.3 & 51    & 195   & \textbf{3,836} & 84.6  & 1.3 & 65.89 \\
      LoHi-SemDiv  & 3.3 & 120   & 195   & \textbf{3,906} & 85.8  & 1.3 & \textbf{66.81} \\
      \bottomrule
      \end{tabular}
      }
  \end{minipage}
  \vspace{-\intextsep}
  \end{figure}

\textbf{\method{} resolves the trade-off between resolution-sensitive (R) and insensitive ($\neg$R) tasks.}
As demonstrated in \lesson{2} and \cref{tab:rsplit}, R and $\neg$R tasks inherently demand divergent configurations. By leveraging complementary representations from both the Lo-V and Hi-I streams, all three proposed \method{} variants simultaneously outperform the single-configuration baselines on both subsets. This consistent gain stems from maintaining dual operating points across the video and image pathways, rather than relying on any specific Hi-I selector.

\textbf{End-to-End Inference Efficiency.} 
\cref{tab:ttft_videomme} profiles the Time-to-First-Token (TTFT) wall time. Unlike prior baselines that decode 256 frames, \method{} decodes only 128 frames (\lesson{3}), immediately reducing front-end latency from 5.4\,s to 3.3\,s. The Key-frame row  represents AKS\cite{tang2025adaptive}, BOLT\cite{liu2025bolt} and CLIP-TopK. More crucially, token pruning methods (e.g., VisionZip, FlashVid) exhibit a severe Vision Tower bottleneck (464.3 TFLOPs; 2{,}486\,ms) because they encode native-resolution frames before pruning. 
By avoiding dense native-resolution encoding, \method{}
reduces ViT latency by approximately $12\times$
(from 2,486 to 195\,ms), with a measured GPU memory
footprint of 9.6\,GB.
Compared with the dense low-resolution baseline,
\method{}-SemDiv improves VideoMME accuracy from 64.44\%
to 66.81\% while using half as many sampled frames
(128 vs.\ 256) and reducing TTFT from 5,963 to 3,906\,ms.
This result illustrates how joint allocation improves
both accuracy and efficiency.

\subsection{Rethinking Token Pruning for Video-VLMs}
\paragraph{Token pruning falls short of simple low-resolution baselines.} \begin{wraptable}[9]{r}{0.43\linewidth} 
    \centering
    \vspace{-1cm}
    \caption{Token pruning vs Resize.} 
    \vspace{-5pt}
    \label{tab:pruning_vs_resize}
    \setlength{\tabcolsep}{2pt}
    \small
    \resizebox{\linewidth}{!}{
    \begin{tabular}{l c c c c c}
    \toprule
    Method & Setting & VideoMME & LVBench  & MLVU  & Avg.\\
    \midrule
    Qwen3-VL-4B & 16F$@1.0$ & 57.78 & 38.61 & 58.89 & 51.76 \\
    \midrule
    \rowcolor{mygray}
    \multicolumn{6}{l}{\emph{$25\%$ default budget ($\sim$1{,}440 tokens)}} \\
    VisionZip \cite{yang2025visionzip}            & 16F$@1.0$  & 55.56 & 34.47 & 53.75 & 47.93 \\
    MMTok \cite{dong2026mmtok}                & 16F$@1.0$  & 55.19 & 34.80 & 54.82 & 48.27 \\
    FlashVid \cite{fan2026flashvid} & 16F$@1.0$  & 55.11 & 35.44 & 54.54 & 48.36 \\
    Resize                                & 16F$@0.5$  & \textbf{56.26} & \textbf{35.51} & \textbf{57.95} & \textbf{49.91} \\
    \midrule
    \rowcolor{mygray}
    \multicolumn{6}{l}{\emph{$50\%$ default budget ($\sim$2{,}880 tokens)}} \\
    VisionZip \cite{yang2025visionzip}            & 128F$@1.0$ & 54.44 & 35.51   & 58.99 & 49.65 \\
    MMTok \cite{dong2026mmtok}                & 128F$@1.0$ & 59.56 & 36.86 & 62.24 & 52.89 \\
    FlashVid \cite{fan2026flashvid} & 128F$@1.0$ & 58.19 & 35.77 & 60.67 & 51.54 \\
    Resize                                & 128F$@0.25$ & \textbf{63.00} & \textbf{39.25} & \textbf{66.34} & \textbf{56.20} \\
    \bottomrule
    \end{tabular}
    \vspace{-10cm}
    }
\end{wraptable}

To analyze the limitations of token pruning exhibited in \cref{tab:main},
we further benchmark pruning methods against a naive resize baseline under fixed token budgets (25\% and 50\%; \cref{tab:pruning_vs_resize}), including MMTok~\cite{dong2026mmtok}, which is explicitly optimized for extreme low-retention regimes.
Despite feeding identical frame counts and tokens to the LLM, their pipelines differ fundamentally: resizing downsamples frames upfront (drastically reducing Vision Tower FLOPs), whereas pruning relies on costly native-resolution processing before discarding tokens. Strikingly, simple resizing consistently matches or outperforms all dedicated pruning methods. This exposes a fundamental inefficiency in current paradigms, proving that resizing is both computationally cheaper and empirically superior.

\textbf{Spatiotemporal consistency is critical for Video-VLMs.} Although effective for static images~\cite{yang2025visionzip,dong2026mmtok}, token pruning degrades temporal position modeling in videos. Discarding irregular patches across frames destroys both spatial continuity and temporal consistency~\cite{fan2026flashvid}. Consequently, preserving a complete, lower-resolution global context is strictly more critical for dynamic video comprehension than retaining fragmented, high-resolution details.

\subsection{Ablations}
\label{sec:ablations}

We ablate the three core design axes of \method{} on VideoMME using Qwen3-VL-4B: the number of high-resolution images (Hi-I$_K$), the high resolution image selection strategy, and the necessity of dual pathways.

\begin{table*}[h]
\centering
\vspace{-5pt}
\begin{minipage}[t]{0.48\linewidth}
\centering
\caption{{K of Hi-I ablation across selectors.}}
\vspace{-5pt}
\label{tab:k_ablation}
\setlength{\tabcolsep}{5pt}
\small
\begin{tabular}{l c c c}
\toprule
Selector & $K{=}2$ & $K{=}4$ & $K{=}8$ \\
\midrule
LoHi-Uniform       & 65.44          & 65.52          & 65.78 \\
LoHi-Anchor        & 65.22          & 65.85          & 65.89 \\
LoHi-CLIP Top-$K$  & 65.37          & 66.00          & 66.44 \\
LoHi-SemDiv        & \textbf{65.59} & \textbf{66.22} & \textbf{66.81} \\
\bottomrule
\end{tabular}

\vspace{-10pt}
\end{minipage}
\begin{minipage}[t]{0.48\linewidth}
\providecommand{\cmark}{\textcolor{green!55!black}{\ding{51}}}
\providecommand{\xmark}{\textcolor{red!70!black}{\ding{55}}}
\centering
\caption{{Lo-V / Hi-I necessity ablation.}}
\vspace{-5pt}
\label{tab:lohi_decomposition}
\setlength{\tabcolsep}{4.5pt}
\small
\resizebox{0.83\linewidth}{!}{
\begin{tabular}{c c | c | c c}
\toprule
Lo-V & Hi-I & Tokens & Uniform & SemDiv \\
\midrule
256F   &   -  &5{,}760    & 64.44 & --    \\
- &   8   &2{,}880  & 55.19 & 60.85 \\
- &   16  &5{,}760  & 60.41 & 63.74 \\
\midrule
128F &  8           &5{,}760  & 65.78 & 66.81 \\
\bottomrule
\end{tabular}
\vspace{-10pt}
}
\end{minipage}
\vspace{-5pt}
\end{table*}

\textbf{Robustness across frame counts and selectors.}
To evaluate the architectural stability of \method{}, we investigate its sensitivity to the high-resolution frame count ($K\in\{2,4,8\}$) and four training-free selection strategies (\cref{tab:k_ablation}). All 12 configurations achieve 65.22--66.81\% accuracy on VideoMME, exceeding the 128-frame Lo-V-only baseline (63.00\%). This consistency suggests that the gains are not specific to a single selector or Hi-I budget. SemDiv further improves over Uniform at every budget, highlighting the benefit of query-aware, diversity-aware selection. Beyond these small Hi-I budgets, we also evaluate LoHi on EgoLongQA~\citep{egowearbench2026} with Qwen3.5-27B, where it matches the accuracy of a higher-resolution single-stream baseline using 29\% fewer visual tokens (\cref{sec:app_larger_budgets}).

\textbf{Lo-V and Hi-I are strictly complementary.}
To prove the necessity of both pathways (Lo-V and Hi-I), we isolate their individual contributions under a strictly controlled budget of 5,760 tokens (\cref{tab:lohi_decomposition}). Relying exclusively on dense low-resolution video (256 Lo-V frames) or sparse high-resolution images (Hi-I$_{K=16}$, SemDiv) yields suboptimal performance at 64.44\% and 63.74\%, respectively. In stark contrast, integrating both pathways—allocating tokens to Lo-V with 128 frames and Hi-I with 8 images—boosts accuracy to 66.81\%. This confirms our core hypothesis: dense temporal context (Lo-V) and precise spatial clarity (Hi-I) provide orthogonal, non-replaceable benefits.

\begin{figure}[h]
\centering
\vspace{-2pt}
\includegraphics[width=\linewidth]{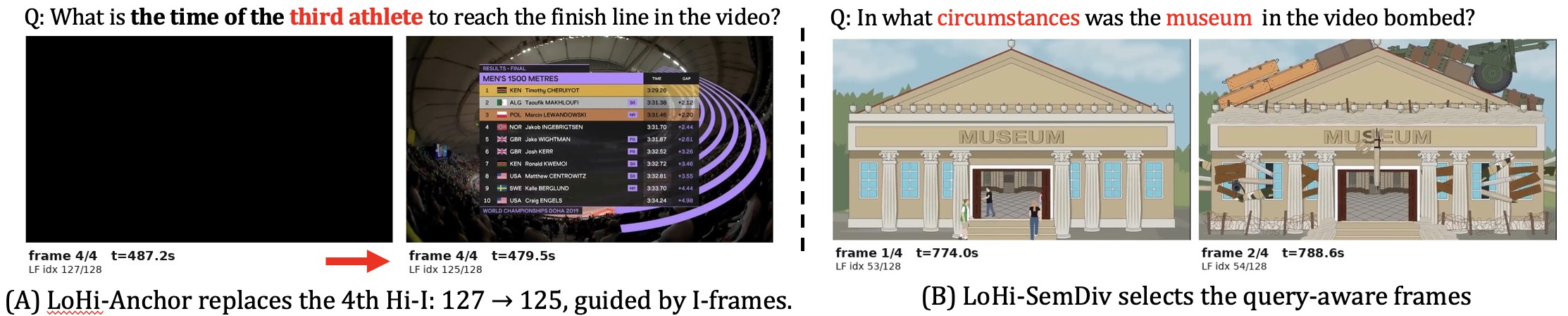}%
\vspace{-0.1cm}
\caption{Visualization of selected high-resolution images by our proposed Hi-I Selector.}
\vspace{-0.4cm}
\label{fig:vis_lohi}
\end{figure}

\textbf{Qualitative analysis.} \cref{fig:vis_lohi} visualizes Hi-I frames chosen by the two selectors. 
(a) \method{}-Anchor replaces a black frame with the Lo-V grid frame closest to the nearest I-frame, recovering the scoreboard.
(b) \method{}-SemDiv picks two query-relevant frames spanning the critical state transition of the museum before and after the bombing. More cases are provided in~\cref{sec:visualization}.
\vspace{-5pt}

\vspace{-5pt}

\section{Conclusion}
\label{sec:conclusion}
We revisit long-video VLM efficiency as a joint allocation
problem over frame count, per-frame resolution, and front-end
latency. This perspective motivates \method{}, a training-free, single-pass framework combining dense low-resolution video with sparse high-resolution images. At the same visual-token budget, \method{} improves average accuracy by 10.6\% over the native-resolution baseline and reduces front-end latency by up to $7\times$ on hour-scale clips. Its plug-and-play selectors support different cost--accuracy trade-offs. Future work can explore richer selectors and adaptive allocation.

\bibliographystyle{plainnat}
\bibliography{refs}

\medskip

\appendix

\newpage
\appendix
\section{Extra Method and Experiment Details}
\label{sec:app_method_extra}

This appendix supplements \cref{sec:method}. It gives the formal definition of \method{}-Anchor, the algorithm and complexity of \method{}-SemDiv, and two exploratory variants that extend the SemDiv selector: \method{}-SemDiv-Anchor, which combines the codec and semantic signals, and \method{}-SemDiv++, which conditions the selection on the Lo-V stream.

\subsection{LoHi-Anchor}
\label{sec:app_anchor}

\method{}-Anchor starts from the $K$ uniformly spaced indices $\{u_1,\dots,u_K\}$ used by \method{}-Uniform. For each position, it considers the five nearest I-frame candidates to avoid duplicate selections, then maps the selected I-frame back to the nearest Lo-V grid index:

\begin{equation}
h_m = \arg\min_{i \in \{1,\dots,N\}} |p_i-j_m|,
\qquad
\mathcal{H}_{\text{Anchor}} = \{h_m\}_{m=1}^{K},
\label{eq:anchor_appendix}
\end{equation}
where $p_i$ is the original-video frame index of the $i$-th Lo-V grid frame, and $j_m$ is the I-frame index selected for $u_m$. I-frame indices are retrieved using \texttt{decord.VideoReader.get\_key\_indices()} (\cref{sec:app_codec_extract}). No temporal-distance threshold is applied. If keyframe retrieval fails or returns no keyframes, the selector falls back to \method{}-Uniform.

All Hi-I frames are selected from the same decoded Lo-V candidate pool and resized at high resolution, requiring no additional frame decoding. The selector runs no CLIP forward pass and does not use the~question.

\subsection{LoHi-SemDiv}
\label{sec:app_semdiv_alg}

\cref{alg:semdiv} shows the greedy MAP procedure for the regularized log-determinant objective $g(\mathcal{S})=\log\det(L_\mathcal{S}+\mathbf{I})$ in \cref{eq:semdiv_obj}. At each step, the algorithm picks the frame that increases $g$ the most given the frames already selected. By \Cref{prop:semdiv_greedy}, the resulting set is within a $(1{-}1/e)$ factor of the optimum.

\begin{algorithm}[H]
\caption{LoHi-SemDiv: greedy MAP for the regularized log-det objective.}
\label{alg:semdiv}
\begin{algorithmic}[1]
\STATE \textbf{Input:} kernel $L \in \mathbb{R}^{N \times N}$, budget $K$, objective $g$ (\cref{eq:semdiv_obj}).
\STATE Initialize $\mathcal{S} \leftarrow \emptyset$.
\FOR{$i = 1, \ldots, K$}
    \STATE $s_i \leftarrow \arg\max_{s \in [N] \setminus \mathcal{S}}\, g\bigl(\mathcal{S} \cup \{s\}\bigr)$
    \STATE $\mathcal{S} \leftarrow \mathcal{S} \cup \{s_i\}$
\ENDFOR
\STATE \textbf{return} $\mathcal{S}$ sorted in temporal order.
\end{algorithmic}
\end{algorithm}

With rank-one Cholesky updates on $L_\mathcal{S}+\mathbf{I}$, the procedure runs in $\mathcal{O}(NK^2)$ time. On top of the Lo-V decoding that \method{} performs anyway, SemDiv adds only one CLIP forward pass over the $N$ low-resolution frames. Both costs are negligible compared with VLM prefill.

\subsection{LoHi-SemDiv-Anchor}
\label{sec:app_semdiv_anchor}
\label{app:semdiv_anchor}

\method{}-SemDiv-Anchor combines the two selectors above. It runs the SemDiv greedy search on snapped positions, so the diversity kernel scores exactly the frames that the VLM will receive. At each greedy step, the candidate index is snapped to the nearest Lo-V grid index associated with a nearby I-frame, subject to a radius $t(D,K)$, and the kernel is then evaluated at the snapped position. The radius is measured in Lo-V grid steps and depends on the clip duration $D$ and the budget $K$:
\begin{equation}
t(D, K) =
\begin{cases}
0 & D < 180\,\text{s},\\
\lfloor 0.10\,N/K \rfloor & D \ge 180\,\text{s},
\end{cases}
\label{eq:semdiv_anchor_threshold}
\end{equation}
which gives $t=3$ at $K=4$ and $t=1$ at $K=8$ for $N=128$ on long clips. On short clips, the snap is disabled because moving an index could shift it beyond the short time span that a typical action occupies; the variant then reduces to plain SemDiv. 
To avoid duplicate grid selections, the snap checks the four nearest I-frames in order of distance and accepts the first whose mapped Lo-V grid index lies within the radius and has not been selected; if none qualifies, the index is left unchanged. \cref{alg:semdiv_anchor} shows the full loop.

\begin{algorithm}[H]
\caption{LoHi-SemDiv-Anchor: snap-aware greedy MAP.}
\label{alg:semdiv_anchor}
\begin{algorithmic}[1]
\STATE \textbf{Input:} kernel $L \in \mathbb{R}^{N \times N}$, budget $K$, I-frame set $\mathcal{I}$, duration $D$, threshold $t(D,K)$.
\STATE Initialize $\mathcal{S}_{\text{snap}} \leftarrow \emptyset$.
\FOR{$i = 1, \ldots, K$}
    \STATE $j^\star \leftarrow \arg\max_{j \in [N] \setminus \mathcal{S}_{\text{snap}}} \log\det\bigl(L_{\mathcal{S}_{\text{snap}}\cup\{\tau(j;\,\mathcal{S}_{\text{snap}})\}} + \mathbf{I}\bigr)$
    \STATE $\mathcal{S}_{\text{snap}} \leftarrow \mathcal{S}_{\text{snap}} \cup \{\tau(j^\star;\,\mathcal{S}_{\text{snap}})\}$
\ENDFOR
\STATE \textbf{return} $\mathcal{S}_{\text{snap}}$ sorted in temporal order.
\end{algorithmic}
\end{algorithm}
Here, $\tau(j;S)$ denotes the snapped Lo-V grid index returned by the procedure above, or $j$ if no valid candidate is found.

We set the constants (the $0.10$ fraction, the $180$\,s threshold, and the four snap candidates) by rough reasoning rather than tuning. The procedure stays $\mathcal{O}(NK^2)$ and requires only keyframe-index lookup in addition to the CLIP inference used by SemDiv. On VideoMME (\cref{tab:semdiv_anchor_app}), the variant improves over plain SemDiv by $+0.19\%$ at $K{=}4$ and $+0.08\%$ at $K{=}8$, mostly on medium-length clips, while at $K{=}2$ the two selectors perform on par. Given the small gains, we regard this variant as exploratory.

\begin{table}[h]
\centering
\caption{\method{}-SemDiv-Anchor on VideoMME (Qwen3-VL-4B, $128$F Lo-V, interleaved cue).}
\label{tab:semdiv_anchor_app}
\setlength{\tabcolsep}{6pt}
\small
\begin{tabular}{l c c c c c}
\toprule
Selector & $K$ & Short & Medium & Long & All \\
\midrule
\method{}-SemDiv         & 2 & 75.67 & 65.78 & 55.33 & 65.59 \\
\method{}-SemDiv-Anchor  & 2 & 75.67 & 65.67 & 55.22 & 65.52 \\
\midrule
\method{}-SemDiv         & 4 & 76.22 & 65.22 & 57.22 & 66.22 \\
\method{}-SemDiv-Anchor  & 4 & 76.22 & 66.00 & 57.00 & \textbf{66.41} \\
\midrule
\method{}-SemDiv         & 8 & 77.11 & 65.78 & 57.56 & 66.81 \\
\method{}-SemDiv-Anchor  & 8 & 77.11 & 66.00 & 57.56 & \textbf{66.89} \\
\bottomrule
\end{tabular}
\end{table}

\subsection{LoHi-SemDiv++: Lo-V-Conditioned Hi-I Selection}
\label{sec:app_lo_residual}

SemDiv in \cref{sec:semdiv} rests on a simple premise: the Lo-V base already covers the timeline, so the $K$ Hi-I frames only need to be relevant to the query and different from each other. Its quality--similarity DPP is built from the CLIP features alone and never refers to the Lo-V base. A natural question is whether an explicit reference to Lo-V would help, that is, whether Hi-I should target the visual content that Lo-V does \emph{not} already capture.

\method{}-SemDiv++ does this with a PCA projection. Let $\{e_j^{\ell}\}_{j=1}^{N} \subset \mathbb{R}^d$ be the L2-normalized CLIP embeddings of the $N$ Lo-V frames, and $U_r \in \mathbb{R}^{d \times r}$ the top-$r$ PCA components fitted on them. We project each Hi-I candidate embedding $e_j$ onto the subspace orthogonal to these components and renormalize:
\begin{equation}
\hat{e}_j \;=\; \frac{(\mathbf{I} - U_r U_r^{\top})\, e_j}{\bigl\lVert(\mathbf{I} - U_r U_r^{\top})\, e_j\bigr\rVert_2}.
\label{eq:semdiv_pp_proj}
\end{equation}
The SemDiv kernel in \cref{eq:semdiv_kernel} is then computed on $\{\hat{e}_j\}$ instead of $\{e_j\}$; the quality term and the greedy procedure in \cref{alg:semdiv} stay unchanged.

\method{}-SemDiv++ improves slightly over plain SemDiv (\cref{tab:lo_residual_app_merged}), which suggests that modeling Lo-V as an explicit information base is a promising direction. Since the current gain is small, we keep plain SemDiv as the default and leave a deeper study of Lo-V conditioning to future work.

\begin{table}[h]
\centering
\caption{\method{}-SemDiv++ on Qwen3-VL-4B ($128$F$@0.25$, $K{=}8$).}
\label{tab:lo_residual_app_merged}
\setlength{\tabcolsep}{4pt}
\small
\begin{tabular}{l c c c c c c}
\toprule
 & \multicolumn{4}{c}{VideoMME} & & \\
\cmidrule(lr){2-5}
Variant & Short & Medium & Long & All & MLVU  & LVBench \\
\midrule
\method{}-SemDiv (reference)   & 77.11 & 65.78 & 57.56 & 66.81 & 74.38 & \textbf{45.90} \\
\method{}-SemDiv++ (frame-PCA) & 77.22 & 66.22 & 57.33 & \textbf{66.93} & \textbf{75.02} & 45.77 \\
\midrule
$\Delta$                       & $+$0.11 & $+$0.44 & $-$0.23 & $+$0.12 & $\mathbf{+0.64}$ & $-$0.13 \\
\bottomrule
\end{tabular}
\end{table}

\section{Extra Implementation Details}
\label{sec:app_implementation}

\subsection{Hardware and software configuration}
\label{sec:app_hardware}

All measurements use a single node with four Intel Xeon Gold 6548Y CPUs and $512$\,GB DDR5 memory, together with
one NVIDIA H100 80\,GB PCIe GPU. Front-end decoding uses \texttt{decord} built against FFmpeg/libavcodec. The GPU NVDEC backend is invoked through the same \texttt{decord} interface. VLM inference runs through HuggingFace \texttt{transformers} 4.51.3 with PyTorch 2.5 in bf16. Models (Qwen3-VL-4B/8B, VideoLLaMA3-7B, Qwen3.5-4B, Qwen2.5-VL-7B, VideoChat3-4B) are pulled from their official checkpoints and rsynced to local NVMe before benchmarking to remove network jitter. All multiple-choice benchmarks are evaluated with \texttt{lmms-eval}\cite{lmms_eval2024}  using single-turn inference and a maximum of eight generated tokens.

\subsection{Front-end decoding latency profiling}
\label{sec:app_decode_profile}

We profile decode wall time per video on the full VideoMME set of $900$ clips ($85.3\%$ at $1280{\times}720$ and $13.1\%$ at $640{\times}360$, all H.264; median duration about $8.1$\,min, 95th percentile about $56$\,min). For each protocol, we record the wall time from constructing the \texttt{VideoReader} to the return of the final \texttt{get\_batch} call, excluding any later preprocessing or transfer to the GPU. When reporting latency at the reference durations of $10$, $30$, and $60$ minutes, we average over clips within $\pm 2$\,min of each target, so that clips near a duration boundary do not blur the comparison.

\paragraph{CPU vs.\ GPU NVDEC.}
\label{app:gpu_pipeline}
We compare uniform sampling with two backends on the same hardware: CPU \texttt{decord} and GPU NVDEC. \cref{tab:cpu_gpu_decode} reports the mean decode time at each frame count. Decode times differ from those used in~\cref{tab:ttft_videomme}, possibly due to different CPU/GPU workloads across runs, but the relative savings from 256 to 128 frames remain simila. NVDEC is $1.7\times$ slower than CPU \texttt{decord} at $N{=}16$ on the full 900-clip set, and $2.3\times$ slower at $N{=}64$ on a 45-clip subsample. There are two reasons. First, the default video pipeline resizes and normalizes frames on the CPU (through PIL or torchvision), so decoded frames must travel back across PCIe before reaching the VLM, and the GPU decoding hardware sits idle in the meantime. Second, each PCIe transfer adds roughly $0.1$--$0.3$\,s of overhead that a CPU-only path never pays. We therefore report all latency numbers with CPU \texttt{decord}, which is also the default backend in common open-source evaluation stacks (\texttt{lmms-eval}, the VideoLLaMA3 utilities, and the HuggingFace video processors). Our main observation, that front-end decoding dominates VLM compute on long clips, holds for both backends.

\begin{table}[h]
\centering
\caption{Decode wall time on VideoMME with uniform sampling. CPU \texttt{decord} is profiled on the full 900-clip set at every frame count; NVDEC is profiled at $N{=}16$ on the same set and at $N{=}64$ on a 45-clip subsample ($\dagger$).}
\label{tab:cpu_gpu_decode}
\small
\setlength{\tabcolsep}{8pt}
\begin{tabular}{c c c c c c}
\toprule
Frames & $N_\text{CPU}$ & CPU mean (s) & $N_\text{GPU}$ & GPU mean (s) & GPU\,/\,CPU \\
\midrule
16   & 900 & 0.60 & 900       & 1.04          & $1.73\times$ \\
64   & 900 & 1.47 & $45^\dagger$ & $3.38^\dagger$ & -\\
128  & 900 & 2.45 & ---       & ---           & ---          \\
256  & 900 & 3.94 & ---       & ---           & ---          \\
\bottomrule
\end{tabular}
\end{table}

\paragraph{Decode at native resolution, resize downstream.}
\label{app:why_native_decode}
Our pipeline decodes every requested frame at native resolution and resizes it afterwards with \texttt{smart\_resize}; the Lo-V and Hi-I streams share this single decode pass and differ only in the resize branch. This design costs little, because H.264 decode time depends on the encoded content (the macroblock and motion-vector count at native resolution) rather than on the output resolution. In our profile, IDCT and motion compensation take roughly $70\%$ of the decode wall time, PCIe and memory copies take about $25\%$, and the resize step takes only about $5\%$. Two options appear to bypass the full decode but do not. The \texttt{width}\,/\,\texttt{height} arguments of \texttt{decord} still run the full IDCT and skip only the resize step, and the hardware downscaler of NVDEC suffers from the same random-seek stalls as the GPU pipeline above. Either option saves at most about $5\%$ of decode time at $256$ frames. The decode workload of \method{} is therefore one shared native-resolution pass that produces the $128$ Lo-V frames together with the $K$ Hi-I frames.

\subsection{TTFT pipeline profile}
\label{sec:app_ttft_protocol}
\cref{tab:ttft_videomme} breaks the TTFT into four parts: decode, selector, ViT, and LLM prefilling. All numbers are measured on a dedicated H100 80\,GB PCIe node. Each component is the median wall-clock latency over $100$ runs after $30$ warm-up runs, with \texttt{torch.cuda.synchronize()} called between samples. To compare methods at the same token budget, we decode $256$ uniformly sampled frames once, store them in a fixed tensor, and reuse this tensor across all runs; the methods then differ only in how they transform these frames (subsampling, CLIP scoring, or token pruning) and in the resulting ViT and LLM prefill shapes. Decode time is not re-measured on this path, since the fixed tensor skips the codec; 
we instead use CPU \texttt{decord} timings measured on the TTFT node ($5.4$\,s for $256$ native frames, $3.3$\,s for $128$). These are higher than the full-set profile in~\cref{tab:cpu_gpu_decode} because of different machine load, but the 256-to-128 ratio is similar (1.64× vs. 1.61×). For the keyframe baselines (CLIP-TopK, AKS, BOLT), the selector latency comes from a shared CLIP-encode-and-select pipeline, about $130$\,ms for $256$ frames at $720{\times}1280$ with CLIP-B/32 on the H100. SemDiv encodes only the $128$ Lo-V candidates, about $65$\,ms, and its greedy DPP solve adds a negligible amount. ViT and LLM prefill are measured directly on Qwen3-VL-4B with the token shape of each method.

\subsection{Codec metadata extraction}
\label{sec:app_codec_extract}

We obtain the keyframe index set $\mathcal{I}$ using \texttt{decord.VideoReader.get\_key\_indices()} and use these indices to guide \method{}-Anchor (\cref{eq:anchor_appendix}). The index lookup requires no additional pixel decoding. Each selected index is mapped to the nearest Lo-V grid frame, so Hi-I frames reuse the decoded candidate pool. VideoReader initialization is included in the reported front-end latency.

\section{Extra Experiments and Analysis}
\label{sec:app_extra_experiments}

\subsection{LoHi on Qwen3-VL-8B}
\label{sec:app_lohi_qwen8b}

To check that the cross-resolution decomposition transfers across model scales, we apply \method{} to Qwen3-VL-8B without retuning any hyperparameter. We keep the same three benchmarks and the same $128$F Lo-V at $r_\ell{=}0.25$, but reduce the Hi-I budget to $K{=}4$, which lowers the total token budget from $\sim$$5{,}760$ to $\sim$$4{,}320$.

All three LoHi variants outperform the prior efficiency
baselines on average while using approximately 25\%
fewer tokens (\cref{tab:transfer}). \method{}-Anchor reaches $67.70\%$ on VideoMME, and \method{}-SemDiv reaches $75.34\%$ on MLVU and $46.80\%$ on LVBench, all above the strongest prior baseline despite the smaller token budget. 
We draw two conclusions. The decomposition transfers to the larger backbone. Moreover, once the backbone has accuracy headroom, $K$ becomes one more efficiency lever: it can be reduced to save tokens while the two-stream design still outperforms Low-Res-Base.

\begin{table}[htbp]
    \centering
    \caption{\textbf{Cross-backbone transfer of LoHi}. All LoHi rows use a $128$F Lo-V at $r_\ell{=}0.25$ paired with $K{=}4$ Hi-I frames at $r_h{=}1.0$. Avg.\ is the mean over VideoMME-All, MLVU, and LVBench.}
    \label{tab:transfer}
    \setlength{\tabcolsep}{4.5pt}
    \resizebox{\linewidth}{!}{%
    \begin{tabular}{l c c c c c c c c c c}
    \toprule
    \multirow{2}{*}{Method} & \#Frames & Setting & Tokens & \multicolumn{4}{c}{VideoMME (\%)} & MLVU & LVBench & Avg. \\
    \cmidrule(lr){5-8}
     & (Decoded) & $N$F${@}r$ & (Budget) & Short & Med. & Long & All & (\%) & (\%) & (\%) \\
    \midrule
    \rowcolor{mygray}
    \multicolumn{11}{l}{\emph{Qwen3-VL-8B}} \\
    Vanilla                       & 16  & $16$F${@}1.0$  & 5{,}760 & 69.89 & 57.56 & 52.67 & 60.04 & 58.95 & 38.22 & 52.40 \\
    CLIP-TopK                     & 256 $\rightarrow$ 16 & $16$F${@}1.0$  & 5{,}760 & 68.44 & 58.44 & 53.33 & 60.07 & 69.73 & 44.09 & 57.97 \\
    AKS \cite{tang2025adaptive}   & 256 $\rightarrow$ 16 & $16$F${@}1.0$  & 5{,}760 & 68.56 & 58.22 & 53.33 & 60.04 & 69.88 & 44.09 & 58.00 \\
    BOLT \cite{liu2025bolt}       & 256 $\rightarrow$ 16 & $16$F${@}1.0$  & 5{,}760 & 72.89 & 59.67 & 54.00 & 62.19 & 69.34 & 43.83 & 58.45 \\
    Low-Res-Base                     & 256 & $256$F${@}0.25$       & 5{,}760 & 78.56 & 64.67 & 57.89 & 67.04 & 71.29 & 44.80 & 61.04 \\
    LoHi-Uniform                  & 128 & $128$F${@}0.25+$Hi-I  & 4{,}320 & 77.89 & 66.89 & 56.00 & 66.93 & 72.02 & 43.32 & 60.76 \\
    LoHi-Anchor                   & 128 & $128$F${@}0.25+$Hi-I  & 4{,}320 & 78.89 & 66.78 & 57.44 & \textbf{67.70} & 72.04 & 44.03 & 61.26 \\
    LoHi-SemDiv       & 128 & $128$F${@}0.25+$Hi-I  & 4{,}320 & 78.56 & 65.11 & 57.67 & 67.11 & \textbf{75.34} & \textbf{46.80} & \textbf{63.08} \\
    \bottomrule
    \end{tabular}%
    }
\end{table}

\subsection{Generalization to Qwen3.5}
\label{sec:app_qwen35}

Qwen3.5 is a recently released native multimodal model. We repeat the two main experiments of \cref{sec:experiments} on Qwen3.5-4B, again without retuning any selector hyperparameter, to check that the cross-resolution decomposition depends on the input format rather than on a specific model family.

\textbf{Setup.} By default, the chat template of Qwen3.5 prepends an implicit thinking prefix. We disable thinking and use greedy decoding, matching the inference protocol of \cref{sec:experiments}.

\paragraph{The fixed-budget trade-off transfers.} At the same token budget, we compare the few-frame, high-resolution setting ($16$F$@1.0$) with the many-frame, low-resolution setting ($256$F$@0.25$) on the three benchmarks. The dense low-resolution recipe wins on every benchmark (\cref{tab:qwen35_iso_app}).

\begin{table}[h]
\centering
\caption{Frame--resolution trade-off on Qwen3.5-4B at the same token budget.}
\label{tab:qwen35_iso_app}
\setlength{\tabcolsep}{10pt}
\small
\begin{tabular}{l c c c}
\toprule
Setting & VideoMME & MLVU & LVBench \\
\midrule
\rowcolor{mygray}
\emph{Qwen3.5-4B} & & & \\
$16$F$@1.0$   & 60.74 & 60.48 & 38.54 \\
$256$F$@0.25$ & \textbf{67.41} & \textbf{69.87} & \textbf{41.45} \\
\midrule
$\Delta$      & $+6.67$ & $+9.39$ & $+2.91$ \\
\bottomrule
\end{tabular}
\end{table}

\paragraph{LoHi transfers.} We then run \method{} on Qwen3.5-4B with a $128$F Lo-V at $r_\ell{=}0.25$ and $K{=}8$ Hi-I frames, and compare the three selectors (\cref{tab:qwen35_lohi_app}). SemDiv is the best selector on all three benchmarks.It also improves over the $256$F$@0.25$ Low-Res-Base on every benchmark at the same token budget, with the largest gain on LVBench (+5.23\%), whose long clips leave the most room for diversity-aware selection. SemDiv also Both findings therefore carry over to this newer natively multimodal backbone without any modification.

\begin{table}[h]
\centering
\caption{\method{} selectors on Qwen3.5-4B ($128$F Lo-V at $r_\ell{=}0.25$, $K{=}8$ Hi-I).}
\label{tab:qwen35_lohi_app}
\setlength{\tabcolsep}{8pt}
\small
\begin{tabular}{l c c c}
\toprule
Configuration & VideoMME & MLVU & LVBench \\
\midrule
\rowcolor{mygray}
\emph{Qwen3.5-4B} & & & \\
$128$F$@0.25$          & 64.93 & 66.80 & 39.51 \\
$256$F$@0.25$          & 67.41 & 69.87 & 41.45 \\
\midrule
\method{}-Uniform                  & 67.52 & 67.48 & 44.61 \\
\method{}-Anchor                   & 67.70 & 66.70 & 44.74 \\
\method{}-SemDiv                   & \textbf{68.15} & \textbf{70.42} & \textbf{46.68} \\
\bottomrule
\end{tabular}
\end{table}

\subsection{Generalization to Additional VLM Backbones}
\label{sec:more_vlms}
To further evaluate the generality of LoHi, we extend our
experiments to three additional VLMs: Qwen2.5-VL-7B\cite{bai2025qwen2},
VideoChat3-4B\cite{li2026videochat3}, and VideoLLaMA3-7B\cite{zhang2025videollama}.

\noindent Across these backbones, the results further support
our three lessons. \lesson{1}: the dense low-resolution base consistently outperforms the 16-frame vanilla setting at the same token budget. \lesson{2}: LoHi-SemDiv improves over the dense base in seven of nine model--benchmark combinations, with small decreases in the remaining two ($-0.26\%$ and $-0.13\%$). \lesson{3}: LoHi requires only 128 sampled frames to be decoded instead of 256, reducing the front-end decoding workload.

\begin{table}[h]\centering\small
\caption{Generalization to additional VLM backbones.
Both LoHi variants use 128 Lo-V frames at $r_\ell=0.25$
and Hi-I frames at $r_h=1.0$. $\dagger$For VideoLLaMA3, MLVU scores use the output normalization described in~\cref{sec:app_vl3_mlvu_norm}.}
\label{tab:additional_backbones}
\setlength{\tabcolsep}{4pt}
\begin{tabular}{lccc}
\toprule
Method & VideoMME & MLVU & LVBench \\
\midrule
\rowcolor{mygray}
\emph{Qwen2.5-VL-7B} & & & \\
Vanilla 16F@1.0        & 58.78 & 58.00 & 37.83 \\
Low-Res-Base 256F@0.25 & 63.78 & 64.77 & 41.06 \\
LoHi-Uniform $K{=}8$   & 63.26 & 63.80 & 40.35 \\
LoHi-SemDiv $K{=}8$    & \textbf{64.41} & \textbf{68.03} & \textbf{43.51} \\
\midrule
\rowcolor{mygray}
\emph{VideoChat3-4B} & & & \\
Vanilla 16F@1.0        & 59.78 & 63.43 & 40.09 \\
Low-Res-Base 256F@0.25 & 64.81 & 70.29 & \textbf{46.29} \\
LoHi-Uniform $K{=}8$   & 63.56 & 69.18 & 42.80 \\
LoHi-SemDiv $K{=}8$    & \textbf{64.96} & \textbf{73.00} & 46.03 \\
\midrule
\rowcolor{mygray}
\emph{VideoLLaMA3-7B$\dagger$ } & & & \\
Vanilla 16F@1.0        & 46.81 & 52.38 & 32.09 \\
Low-Res-Base 256F@0.25 & 50.04 & 57.73 & 36.99 \\
LoHi-Uniform $K{=}8$   & \textbf{51.37} & 58.14 & 35.57 \\
LoHi-SemDiv $K{=}8$    & 50.67 & \textbf{58.59} & 36.86 \\
\bottomrule
\end{tabular}
\end{table}

\subsection{Pushing the Lo-V to $N{=}256$}
\label{sec:app_lohi_256}

The main paper fixes the Lo-V at $N{=}128$ so that all methods compare at the same token budget. Here we relax this constraint: we keep $r_\ell{=}0.25$ and $r_h{=}1.0$, increase the Lo-V to $N{=}256$, and add $K{=}4$ Hi-I frames. \cref{tab:lohi_256} reports VideoMME results on Qwen3-VL-4B. The results extend both lessons. Densifying the Lo-V improves accuracy by supplying richer temporal coverage (\lesson{1}), and adding Hi-I frames on top still brings a clear gain, since the high-resolution evidence they supply (\lesson{2}) is not something more low-resolution frames can provide. The two streams therefore remain complementary as the Lo-V grows denser.

\begin{table}[h]
\centering
\caption{Pushing the Lo-V to $N{=}256$ on VideoMME (Qwen3-VL-4B). All rows use $r_\ell{=}0.25$ for Lo-V and $r_h{=}1.0$ for Hi-I.}
\label{tab:lohi_256}
\setlength{\tabcolsep}{6pt}
\small
\begin{tabular}{l c c c c c c}
\toprule
Configuration & Lo-V & Hi-I & Short & Medium & Long & All \\
\midrule
\rowcolor{mygray}
\emph{Qwen3-VL-4B} & & & &&&\\
$256$F$@0.25$         & $256$F & ---     & 75.78 & 62.33 & 55.22 & 64.44 \\
\midrule
\method{}-Uniform              & $128$F & $K{=}8$ & 75.67 & 66.44 & 55.22 & 65.78 \\
\method{}-SemDiv               & $128$F & $K{=}8$ & 77.11 & 65.78 & 57.56 & \textbf{66.81} \\
\midrule
\method{}-Uniform  & $256$F & $K{=}4$ & 77.67 & 65.33 & 56.78 & 66.59 \\
\method{}-Anchor   & $256$F & $K{=}4$ & 77.00 & 65.89 & 57.78 & 66.89 \\
\method{}-SemDiv   & $256$F & $K{=}4$ & 76.78 & 66.44 & 58.11 & \textbf{67.11} \\
\bottomrule
\end{tabular}
\end{table}

\subsection{Evaluation under Larger Token Budgets}
\label{sec:app_larger_budgets}

To cover a substantially wider budget range, we evaluate Qwen3.5-27B on EgoLongQA, the long-video QA task of EgoWearBench~\citep{egowearbench2026}. We use 700 egocentric clips of approximately 10 minutes each. To examine the token allocation motivated by \lesson{1} and \lesson{2} in this regime, we relax the decoded-frame constraint and use 512 sampled frames for all configurations. For LoHi, we fix the Lo-V resolution at $r=0.25$ and increase the Hi-I budget from $K=0$ to $K=128$, scaling the visual-token count from approximately 25K to 117K. We use SemDiv for Hi-I selection. We also compare against a single-stream baseline at $r=0.5$. Results are reported in \cref{tab:budget}.

\begin{table}[ht]
\centering
\small
\caption{Evaluation under larger visual-token budgets on EgoLongQA with Qwen3.5-27B. All configurations use 512 sampled frames.}
\label{tab:budget}
\setlength{\tabcolsep}{6pt}
\begin{tabular}{lcc}
\toprule
Lo-V $+$ Hi-I & Visual tokens & Acc.\ (\%) \\
\midrule
\rowcolor{mygray}
\emph{Qwen3.5-27B} & & \\
512F@0.25, $K=0$   & 25,088  & 82.14 \\
512F@0.25, $K=8$   & 30,848  & 82.86 \\
512F@0.25, $K=16$  & 36,608  & 83.71 \\
512F@0.25, $K=32$  & 48,128  & 85.00 \\
512F@0.25, $K=64$  & \textbf{71,168} & \textbf{85.86} \\
512F@0.25, $K=128$ & 117,248 & \textbf{87.57} \\
\midrule
512F@0.5, $K=0$    & 100,352 & \textbf{85.86} \\
\bottomrule
\end{tabular}
\end{table}

\noindent\textbf{(i) Adding more Hi-I frames consistently helps.}
Accuracy improves monotonically from 82.14\% to 87.57\% as $K$ increases from 0 to 128.
\textbf{(ii) At matched accuracy, LoHi uses less visual computation.}
512F@0.25 with $K=64$ attains the same accuracy as 512F@0.5 (85.86\%) while using \textbf{29\% fewer visual tokens} (71K vs.\ 100K) and \textbf{29\% fewer ViT patches} (285K vs.\ 401K).
These results further support our view of long-video efficiency as a joint allocation problem.

\subsection{Adaptive cost control via multi-turn triggering}
\label{sec:app_multiturn}

A benefit of the decoupled two-stream design is that it naturally supports multi-turn inference: the Lo-V base can answer first, and the Hi-I frames can be added in a second pass only when needed. Since the Lo-V base alone already answers most questions correctly (\cref{tab:multiturn_app}), this offers a chance to reduce vision tokens and prefill latency further. We explore this with the entropy-triggered mechanism in \cref{eq:multiturn}: a question enters the second pass only when the entropy of its Lo-V answer distribution exceeds a threshold. We evaluate on VideoMME with Qwen3-VL-4B and a $128$F Lo-V, and tune the threshold on a held-out $10\%$ split. As \cref{tab:multiturn_app} shows, only about a third of the questions enter the second pass, while the accuracy nearly matches always running Hi-I. The two-stream design thus supports adaptive inference well. The threshold is tuned per benchmark, so we regard this as an exploratory validation; threshold selection without benchmark-specific tuning is left to future work.

\begin{table}[h]
\centering
\caption{Two-pass triggers on VideoMME (Qwen3-VL-4B, $128$F Lo-V).}
\label{tab:multiturn_app}
\small
\setlength{\tabcolsep}{8pt}
\begin{tabular}{l c c c}
\toprule
Trigger & 2nd-pass rate & Acc.\,(\%) & $\Delta$ vs.\ Lo \\
\midrule
Lo-V only                          & $0\%$  & 63.00 & 0.00 \\
Entropy ($\theta$ tuned)           & $37\%$ & {65.71} & $+2.71$ \\
Always-Hi                        & $100\%$ & 65.78 & $+2.78$ \\
\bottomrule
\end{tabular}
\end{table}

\subsection{Answer-Format Normalization for VideoLLaMA3 on MLVU}
\label{sec:app_vl3_mlvu_norm}

This is a formatting issue in answer extraction, not an accuracy effect. MLVU expects answers ending in \texttt{``A)''}. VideoLLaMA3-7B instead writes \texttt{``A.''} in about $30$--$38\%$ of its outputs, and the MLVU extractor then falls back to a literal string comparison that fails to match the gold letter. We fix this by stripping the trailing period whenever the prediction is exactly two characters of the form \texttt{``X.''} with $X \in \{$A,B,C,D$\}$. Every output without a closing parenthesis in the three runs has exactly this form, so the rule covers all such cases. VideoMME and LVBench use a generic \texttt{[ABCD]} regex and are not affected. The normalization raises each VideoLLaMA3 MLVU number by $16$--$17\%$ and does not change the trend across frame counts (\cref{tab:vl3_mlvu_norm}). In \cref{tab:obs1_transfer}, the MLVU scores for VideoLLaMA3
are reported before answer-format normalization and marked
with $\dagger$.

\begin{table}[h]
\centering
\caption{VideoLLaMA3-7B on the MLVU dev split, with and without output normalization. ``No-paren rate'' is the fraction of outputs lacking a closing parenthesis.}
\label{tab:vl3_mlvu_norm}
\setlength{\tabcolsep}{8pt}
\small
\begin{tabular}{l c c c}
\toprule
Configuration & Raw (\%) & Normalized (\%) & No-paren rate \\
\midrule
$16$F$@1.0$    & 35.88 & 52.38 & $38.5\%$ \\
$64$F$@0.5$    & 38.97 & 55.27 & $30.9\%$ \\
$256$F$@0.25$  & 40.30 & 57.73 & $30.8\%$ \\
\midrule
$\Delta$ ($256$F\,$-\,16$F) & $+4.42$ & $+5.35$ & --- \\
\bottomrule
\end{tabular}
\end{table}

\subsection{The inter-frame token compressor in VideoLLaMA3 removes almost no tokens}
\label{sec:app_vl3_difffp}

VideoLLaMA3-7B compresses visual tokens with a module called DiffFP. For each patch, DiffFP computes the pixel difference between adjacent frames and drops the corresponding token when this difference falls below a threshold. To check how much the module actually removes, we measure its keep ratio, defined as the number of tokens after DiffFP divided by the number before. We evaluate $45$ configurations in total, covering five frame counts ($16$ to $256$), three resolution scales ($1.0$, $0.5$, and $0.25$), and three benchmarks, with $60$ videos sampled per configuration.

The result is consistent. The mean keep ratio of every configuration stays above $0.959$ (\cref{tab:vl3_difffp}), so on average DiffFP removes at most $\sim$$4\%$ of tokens in every long-video setting we test. Individual videos can go lower (down to $0.639$ on MLVU), but such cases are rare. The cause is saturation. When many frames are sampled at reduced resolution, each merged patch covers a large image region, and almost any motion between frames pushes its pixel difference above the threshold, so the mask keeps nearly every token. For all token-budget calculations in this paper, we therefore treat VideoLLaMA3 as having no inter-frame compression.

\begin{table}[h]
\centering
\caption{Keep ratio of DiffFP in VideoLLaMA3-7B, $45$ configurations, $60$ videos each. The lowest mean is $0.959$ (VideoMME, $256$F$@1.0$).}
\label{tab:vl3_difffp}
\small
\setlength{\tabcolsep}{8pt}
\begin{tabular}{l c c c}
\toprule
Benchmark & Mean (range) & Std (max) & Min (per video) \\
\midrule
LVBench   & $0.990$\,--\,$0.998$ & $0.032$ & $0.806$ \\
MLVU      & $0.977$\,--\,$0.999$ & $0.069$ & $0.639$ \\
VideoMME  & $0.959$\,--\,$0.991$ & $0.064$ & $0.674$ \\
\bottomrule
\end{tabular}
\end{table}

\subsection{Extra ablations}
\label{sec:app_ablations}

\subsubsection{Prompt template: appended cue vs.\ interleaved cue}
\label{sec:app_prompt_ablation}

The Lo-V and Hi-I streams are joined by a natural-language cue (\cref{sec:method-lohi}). We compare two placements of the cue:

\begin{tcolorbox}[colback=mygray,boxrule=0pt,arc=0pt,left=4pt,right=4pt,top=2pt,bottom=2pt,fontupper=\footnotesize]
\textbf{Appended cue.} \texttt{[Lo-V tokens]\,[Hi-I tokens]} \texttt{``You are watching a video. The video above provides low-resolution temporal context. The $K$ images above are high-resolution views of selected key frames. \{question\}''}
\end{tcolorbox}
\vspace{-2pt}
\begin{tcolorbox}[colback=mygray,boxrule=0pt,arc=0pt,left=4pt,right=4pt,top=2pt,bottom=2pt,fontupper=\footnotesize]
\textbf{Interleaved cue.} \texttt{[Lo-V tokens]} \texttt{``The above low-resolution video provides temporal context. The following $K$ high-resolution images show selected key frames:''} \texttt{[Hi-I tokens]} \texttt{``\{question\}''}
\end{tcolorbox}

In the Appended variant, the model reads the description only \emph{after} it has seen the Hi-I tokens. In the Interleaved variant, the description sits \emph{between} the Lo-V block and the Hi-I block, so the model is told that the following tokens are high-resolution frames before attending to them. \cref{tab:prompt_ablation_app} reports the comparison; all other settings match \cref{tab:k_ablation}.

We find two things. First, \method{} is robust to the prompt wording. Every selector at every $K$ benefits from the cross-resolution decomposition under either prompt, and the gap between the two prompts never exceeds $0.8\%$. Second, a prompt that separates the two modalities more clearly helps further. 
The Interleaved cue adds $+0.44$ to $+0.74\%$ for Uniform and SemDiv.
This suggests that explicitly linking Hi-I images to the video timeline can help. Anchor does not show the same benefit in this ablation.
We use the Interleaved cue by default. Richer delimiters, such as per-frame timestamps or role tokens, may extend this trend further.

\begin{table}[h]
\centering
\caption{Prompt template ablation on VideoMME (Qwen3-VL-4B, $128$F Lo-V at $r{=}0.25$).}
\label{tab:prompt_ablation_app}
\small
\setlength{\tabcolsep}{6pt}
\begin{tabular}{l c c c c}
\toprule
\multirow{2}{*}{Selector} & \multicolumn{2}{c}{$K{=}4$ (\%)} & \multicolumn{2}{c}{$K{=}8$ (\%)} \\
\cmidrule(lr){2-3}\cmidrule(lr){4-5}
                          & Appended & Interleaved & Appended & Interleaved \\
\midrule
\method{}-Uniform         & 64.78 & \textbf{65.52} {\scriptsize$(+0.74)$} & 65.30 & \textbf{65.78} {\scriptsize$(+0.48)$} \\
\method{}-SemDiv          & 65.70 & \textbf{66.22} {\scriptsize$(+0.52)$} & 66.37 & \textbf{66.81} {\scriptsize$(+0.44)$} \\
\method{}-Anchor          & 65.96 & 65.85 {\scriptsize$(-0.11)$}          & 65.85 & 65.89 {\scriptsize$(+0.04)$} \\
\bottomrule
\end{tabular}
\end{table}

\subsubsection{SemDiv sharpening coefficient $\alpha$}
\label{sec:app_alpha_ablation}

The quality term in the SemDiv kernel (\cref{eq:semdiv_kernel}) sharpens the normalized query--frame similarity with an exponent $\alpha$. We sweep $\alpha \in \{1, 5, 10\}$ on VideoMME (Qwen3-VL-4B, $128$F Lo-V at $r{=}0.25$, $K{=}8$, interleaved cue) and obtain $66.67\%$, $\textbf{66.81}\%$, and $66.48\%$. Accuracy ranges from 66.48\% to 66.81\%, suggesting
limited sensitivity to $\alpha$ over the tested range.
We use $\alpha=5$ in all \method{}-SemDiv runs without
per-task tuning.

\subsection{Further Analysis of Hi-I Selection}
\label{sec:app_selection_analysis}

All analyses use Qwen3-VL-4B. We conduct exact McNemar tests and video-level paired bootstrap with 50,000 resamples. The two procedures yield consistent significance conclusions, and the confidence intervals are stable across bootstrap seeds.

\paragraph{Statistical significance.}
As shown in \cref{tab:selection_significance}, SemDiv significantly improves over Uniform on MLVU ($+6.76$\%, $p<10^{-16}$), while the differences on VideoMME and LVBench are not statistically significant.

\begin{table}[h]
\caption{Paired comparison of LoHi-SemDiv and LoHi-Uniform.
Confidence intervals are computed using video-level paired
bootstrap. W/L denotes questions answered correctly only
by SemDiv/Uniform, respectively.}
\label{tab:selection_significance}
\centering
\small
\begin{tabular}{lcccc}
\toprule
SemDiv $-$ Uniform & $\Delta$ (\%) & 95\% CI & McNemar $p$ & W/L \\
\midrule
VideoMME & $+1.04$ & $[-0.15,+2.22]$ & $0.094$ & 144/116 \\
MLVU & $\mathbf{+6.76}$ & $[+5.17,+8.41]$ & $\mathbf{<10^{-16}}$ & 233/86 \\
LVBench & $+1.81$ & $[-0.13,+3.76]$ & $0.063$ & 120/92 \\
\bottomrule
\end{tabular}
\end{table}

\paragraph{Benchmark dependence.}
The same benchmark-dependent pattern appears in CLIP-TopK and AKS: gains over uniform selection are largest on MLVU and small or negative on VideoMME (\cref{tab:selection_benchmark_dependence}). Each comparison uses a matched budget: 16 frames for the keyframe baselines and $K=8$ Hi-I frames for LoHi. The gain magnitudes are not directly comparable, since keyframe methods select the sole visual input, whereas SemDiv selects additional detail on top of Lo-V. The shared pattern shows that benchmark dependence is not unique to SemDiv.

\begin{table}[h]
\caption{Accuracy gains of query-aware selection over
the corresponding uniform baselines at matched budgets.}
\label{tab:selection_benchmark_dependence}
\centering
\small
\begin{tabular}{lccc}
\toprule
Query-aware $-$ uniform selection (\%) & VideoMME & MLVU & LVBench \\
\midrule
CLIP-TopK $-$ Vanilla (16F@1.0) & $-0.6$ & $+10.5$ & $+6.1$ \\
AKS $-$ Vanilla (16F@1.0) & $-0.5$ & $+10.3$ & $+6.1$ \\
LoHi-SemDiv $-$ LoHi-Uniform & $+1.0$ & $+6.8$ & $+1.8$ \\
\bottomrule
\end{tabular}
\end{table}

\paragraph{Task-level benefits.}
Repetitive action counting requires modeling temporal structure across action instances~\citep{hu2022transrac}.MLVU action counting provides a clear example (\cref{tab:selection_action_count}): Uniform falls below the Lo-V-only baseline (29.13\% vs.\ 41.75\%), whereas SemDiv reaches 53.40\%, improving over Uniform by 24.27\% (95\% CI: $[+16.5,+32.0]$). This highlights the importance of Hi-I placement for questions requiring evidence from multiple action instances. The MLVU gain is not driven by this category alone: leave-one-category-out differences range from $+4.93$ to $+8.06$\%, all with $p<10^{-9}$.

\begin{table}[h]
\caption{Effect of Hi-I selection on MLVU action counting
with Qwen3-VL-4B ($n=206$).}
\label{tab:selection_action_count}
\centering
\small
\begin{tabular}{lccc}
\toprule
MLVU action-count & Lo-V only & LoHi-Uniform & LoHi-SemDiv \\
\midrule
Accuracy (\%) & 41.75 & 29.13 & \textbf{53.40} \\
\bottomrule
\end{tabular}
\end{table}

\paragraph{Variation across query types.}
The qualitative examples in
\cref{fig:app_vis_lov,fig:app_vis_semdiv}
illustrate two different needs: resolving fine spatial
details and placing high-resolution frames at the
right moment. We further examine this distinction
through a query-type breakdown.
Within MLVU's ego category, LoHi-Uniform matches Low-Res-Base at 59.66\%, while LoHi-SemDiv reaches 69.03\%. Uniform and SemDiv use the same Lo-V stream and the same Hi-I budget ($K=8$), differing only in which frames are selected. To examine how this benefit varies with the query, we split the 352 ego questions by template. \cref{tab:ego_query_types} reports the five templates with $n\ge20$; the remaining 78 questions belong to templates with at most 17 questions each. Complete results and evaluation logs for all datasets
are available in our GitHub repository, linked from
the project page\footnote{\url{https://sixundong.com/projects/lohi}}.

\begin{table}[ht]
\centering
\small
\caption{Accuracy (\%) by question template within MLVU's ego category, using Qwen3-VL-4B. Base denotes Low-Res-Base; Uniform and SemDiv use $K=8$ Hi-I frames.}
\label{tab:ego_query_types}
\setlength{\tabcolsep}{5pt}
\begin{tabular}{lrccc}
\toprule
Query type & $n$ & Base & Uniform & SemDiv \\
\midrule
Static location (\textit{where was X}) & 111 & 68.47 & 70.27 & \textbf{73.87} \\
Transient action (\textit{put in}) & 48 & 64.58 & 66.67 & \textbf{83.33} \\
Action destination & 45 & 71.11 & 75.56 & \textbf{77.78} \\
Yes/no (\textit{did I \dots}) & 43 & 58.14 & 48.84 & \textbf{65.12} \\
Static entity count & 27 & 18.52 & \textbf{37.04} & \textbf{37.04} \\
\bottomrule
\end{tabular}
\end{table}

Static entity counting improves from 18.52\% to 37.04\% with either selector, whereas transient-action accuracy increases from 66.67\% with Uniform to 83.33\% with SemDiv. These results suggest that high-resolution evidence alone can help some queries, while others benefit substantially from query-aware placement. We leave query-aware adaptive frame--resolution allocation to future work.

\paragraph{Selection overhead.}
SemDiv scores only the 128 Lo-V candidates, adding approximately 70\,ms over Uniform (1.8\% of TTFT). Its total selector-stage latency is 120\,ms, below the 130\,ms of the keyframe-selection baselines (\cref{tab:ttft_videomme}).

\paragraph{Performance across benchmarks and task categories.}
SemDiv is never significantly worse than Uniform on any evaluated benchmark or task category; the decrease on topic reasoning ($-2.66$\%) is not significant. A stratified exact test across all three benchmarks confirms an overall advantage ($p=5\times10^{-13}$). These results and the small additional overhead motivate SemDiv as our default, while Uniform remains an option without semantic-selection overhead.

\subsection{Evaluation with Subtitles}
\label{sec:app_subtitles}

\paragraph{Motivation for the subtitle-free protocol.}
Our main evaluation excludes subtitles to isolate the effect of visual input allocation. To examine how much information subtitles provide, we evaluate Qwen3-VL-4B using only the transcript and question, without visual input. \cref{tab:subtitle_text_only} reports results on VideoMME questions with and without available subtitles; the visual configurations do not use subtitles. With zero visual input, the transcript alone reaches 61.56\% on subtitled questions, close to the 256-frame visual pipeline (63.13\%). Subtitles also consume a substantial context budget. With the Qwen3-VL tokenizer, VideoMME subtitle lengths are heavily right-skewed (\cref{tab:subtitle_lengths}).

\begin{table}[h]
\centering
\small
\caption{Accuracy (\%) of visual-only and text-only inputs
on VideoMME using Qwen3-VL-4B. Subsets are defined by
subtitle availability. Visual inputs exclude subtitles;
text-only inputs contain the transcript and question.}
\label{tab:subtitle_text_only}
\begin{tabular}{lcc}
\toprule
Qwen3-VL-4B & Subtitled (2,232 Qs) & No subtitle (468 Qs) \\
\midrule
16F@1.0       & 56.18 & 65.38 \\
256F@0.25     & 63.13 & 70.73 \\
Text-only     & \textbf{61.56} & -- \\
Question-only & -- & 30.56 \\
\bottomrule
\end{tabular}
\end{table}

\begin{table}[h]
\centering
\caption{VideoMME subtitle lengths measured with the
Qwen3-VL tokenizer, relative to the visual-token
budget of 5,760 tokens.}
\label{tab:subtitle_lengths}
\small
\begin{tabular}{lcc}
\toprule
Subtitle length & Tokens & \% of visual budget \\
\midrule
Median & 1,539 & 27\% \\
Mean   & 3,018 & 52\% \\
P90    & 7,799 & \textbf{135\%} \\
\bottomrule
\end{tabular}
\end{table}

\noindent At the 90th percentile, the transcript alone exceeds our visual-token budget of 5,760 tokens, bringing the combined visual and subtitle input to more than 13,000 tokens. Including subtitles therefore introduces both a substantial source of non-visual information and additional context cost.

\paragraph{Performance with subtitles.}
We evaluate Low-Res-Base and LoHi-SemDiv with and without subtitles on the same subtitled subset ($n=2{,}232$) in \cref{tab:subtitle_comparison}.

\begin{table}
\centering
\caption{Accuracy (\%) with and without subtitles on
the 2,232 subtitle-available questions in VideoMME,
using Qwen3-VL-4B. Reported $p$-values compare
LoHi-SemDiv with Low-Res-Base in each setting.}
\label{tab:subtitle_comparison}
\small
\begin{tabular}{lccc}
\toprule
Subtitled subset & w/o subs & w/ subs & Subtitle gain (\%) \\
\midrule
Low-Res-Base 256F@0.25 & 63.13 & 73.12 & $+9.99$ \\
LoHi-SemDiv $K=8$     & 65.28 & 72.13 & $+6.85$ \\
\midrule
Gap (LoHi $-$ Base) & $+2.15$ & $-0.99$ & -- \\
$p$-value          & 0.011   & 0.117   & -- \\
\bottomrule
\end{tabular}
\end{table}

\noindent Without subtitles, LoHi-SemDiv is significantly more accurate than the dense Low-Res-Base on this subset ($+2.15$\%, $p=0.011$) while using half as many sampled frames. Adding subtitles improves both methods, but the gain is larger for Low-Res-Base ($+9.99$ vs.\ $+6.85$\%), and the resulting difference between the methods is not statistically significant ($-0.99$\%, $p=0.117$). Together with the text-only result, this suggests that subtitles can mask the benefit of improved visual input allocation, motivating our subtitle-free main evaluation. With subtitles, LoHi still uses only 128 sampled frames instead of 256, although its accuracy advantage is no longer observed.

\section{Extended Related Work}
\label{sec:app-related-extended}

\subsection{Codec-Aware and System-Level Video Processing}
\label{sec:app-related-codec-system}

One line of work reduces front-end cost by operating directly on codec-level signals, such as the motion vectors and residuals stored in the compressed bitstream, instead of first decoding to RGB~\citep{sarkar2026cope,zou2026costream}. The idea originates in compressed-domain action recognition~\citep{wu2018compressed}, which uses motion vectors as task features without full decoding. The most recent VLM-side instance, CoPE-VideoLM~\citep{sarkar2026cope}, learns codec-aware token representations and removes the RGB-decode bottleneck end to end. System-level work such as QuickVideo~\citep{schneider2025quickvideo} targets the same problem from another angle, optimizing decoder and prefill throughput at runtime.

These approaches share two properties. They need substantial model-side or system-side training and engineering to make compressed-domain inputs work with downstream VLMs, and their savings come from reducing the cost of decoding each frame. \method{} takes a training-free view one level higher. Instead of redesigning how a single frame is decoded, we observe that long videos need not be decoded densely in the first place: uniformly sampling a fixed budget of $N$ frames already preserves temporal coverage, at a front-end cost that does not grow with duration (\lesson{3}). Uniform sampling also produces evenly spaced timestamps, which unified-stream VLMs need for multimodal RoPE-based temporal modeling. \method{} is therefore complementary to codec-aware tokenizers and runtime systems rather than a replacement. A future system could combine codec-aware decoding with our cross-resolution decomposition to compound the gains.

\subsection{Anyres and Slow--Fast: Two Single-Axis Decompositions}
\label{sec:app-related-anyres-slowfast}

Two paradigms decompose the visual budget along a single axis: anyres tiling along space, and Slow--Fast along time. \method{} differs from both, for opposite reasons: anyres is its enabling foundation, while Slow--Fast conflicts with how modern unified-stream VLMs encode time.

\paragraph{Anyres with token merging makes Lo-V viable.}
Anyres-style methods such as LLaVA-NeXT~\citep{liu2024llavanext} and InternLM-XComposer2-4KHD~\citep{dong2024internlm} split a single high-resolution image into spatial patches, encode the patches independently, and concatenate the results. Modern frontier VLMs (Qwen2.5/3-VL~\citep{bai2025qwen3}, InternVL3.5~\citep{wang2025internvl3_5}) inherit this design and add a vision-side merge layer that aggregates adjacent patches into fewer tokens. As a result, at $r{=}0.25$ on a 720p source, a Qwen3-VL frame occupies only about $20$--$30$ merged tokens (\cref{tab:obs1_transfer} realizes ${\sim}5{,}760$ tokens at $256$ frames). This is exactly the regime in which L1 and L2 operate: the experiments in \cref{sec:obs1,sec:obs2} confirm that moderate downscaling preserves most accuracy, and that the loss concentrates on a small set of resolution-sensitive tasks, which a few high-resolution frames can recover. Tiling is therefore not a competitor to \method{} but the property that makes the low-resolution side of our decomposition cheap and accurate. The two also differ in scope: tiling decomposes within a single static image to improve spatial coverage, while \method{} composes the video and image modalities in one forward pass to balance temporal density and spatial detail under the decode and token constraints of long clips.

\paragraph{Slow--Fast and per-frame variable token allocation.}
The Slow--Fast paradigm~\citep{slowfast} processes video with two pathways at different temporal and spatial rates, originally for action recognition. Video VLMs have imported the same pattern. LLaVA-Video~\citep{zhang2024llava} allocates different token counts to different frames, and query-aware frame selectors such as F2C~\citep{sun2025frames} and Q-frame~\citep{zhang2025q} follow the same spirit: keyframes form a slow path with a larger per-frame budget, and the remaining frames form a fast path with a smaller one. The intent of spending budget where motion or query relevance concentrates is reasonable. The realization, however, conflicts with how unified-stream VLMs encode time. Heterogeneous token counts across frames disrupt the multimodal RoPE position assignment used for temporal reasoning, and recovering this typically requires dedicated training, which rules out plug-and-play deployment on existing backbones.

\paragraph{\method{} keeps the temporal stream uniform and routes detail through the image pathway.}
\method{} keeps a uniform per-frame token allocation in the Lo-V stream, so mRoPE stays consistent across all $N$ frames, and routes additional spatial detail through the image pathway (Hi-I) instead of along the temporal axis. This separation matters because the image pathway is where modern VLMs concentrate their fine-detail capability, such as small-object and text recognition, as seen in DeepSeek-OCR~\citep{wei2025deepseek} and the OCR strength of Qwen3-VL~\citep{bai2025qwen3}. Allocating the high-resolution budget there reuses this learned capability, which is what allows \method{} to be training-free: Lo-V matches the video-side training distribution (a uniform low-resolution timeline), and Hi-I matches the image-side training distribution (high-resolution spatial detail).

\section{Limitations and Outlook}
\label{app:app_discussion}

\paragraph{Limitations.}
We validate \method{} on frontier VLMs (Qwen3-VL-4B/8B, VideoLLaMA3-7B, Qwen3.5-4B, Qwen2.5-VL-7B and VideoChat3-4B) whose dynamic-resolution architectures, such as AnyRes, stay accurate under aggressive downscaling. Older or smaller models such as LLaVA-NeXT~\citep{liu2024llavanext} may degrade more sharply at very low resolutions, and would need a higher base resolution $r_\ell$ or a larger Hi-I budget $K$ to compensate. Our findings therefore apply most directly to modern and future architectures, where low-resolution inputs are highly token-efficient. 
Our training-free integration also relies on a textual cue to bridge the Lo-V and Hi-I streams, introducing mild sensitivity to prompt structure, as shown in the prompt ablation~\cref{sec:app_prompt_ablation}.
In addition, this work focuses on front-end and prefill latency, while the decoding latency of the VLM also matters~\citep{dong2026rethinking}. 
At a matched visual-token budget, LoHi reduces front-end cost without increasing the visual context length used during generation. When fewer visual tokens are desired to reduce prefill and generation costs, the lower-budget results in~\cref{tab:pruning_vs_resize} also support resolution-based allocation. Beyond this controlled setting, agentic keyframe selection can incur additional generation overhead by producing
many reasoning tokens to decide which tools and frames
to use. Existing studies rarely count this thinking time. We leave its evaluation to future work.

\paragraph{Outlook.}
The decoupled two-stream design opens three directions. First, combining with prior efficiency methods. Once the input is split into Lo-V and Hi-I streams, existing keyframe-selection and token-pruning algorithms are no longer competitors; they can serve as drop-in modules that further compress the Hi-I subset. Second, broader adaptive computation. Our entropy-triggered early exit (\cref{sec:app_multiturn}) targets multiple-choice QA, but the same Lo-V-first, Hi-I-on-demand pattern extends naturally to open-ended tasks such as video grounding, summarization, and dialogue. High-resolution inspection could further serve as an on-demand tool for multimodal agents~\citep{wang2025mllm,yin2025livemcp}. Third, reusing intermediate frames. As \lesson{3} shows, decoding a target P-frame forces the decoder to also process the preceding I-frame and P-frames in the same GoP. These frames come for free, and reusing them as extra Hi-I candidates or as motion-magnitude priors could enrich Hi-I selection without increasing the decode budget $N_{\text{dec}}$.

\section{Qualitative Results}
\label{sec:visualization}
We provide additional qualitative examples to illustrate the roles of the Hi-I pathway and the two selectors.

In \cref{fig:app_vis_lov}, the Lo-V-only prediction is wrong because the answer depends on a small detail on the interviewee's chin, which is blurred at low resolution. The selected Hi-I frames restore this detail and lead to the correct answer. The Hi-I pathway is thus most useful when the question needs fine spatial evidence.

\cref{fig:app_vis_anchor} and \cref{fig:app_vis_semdiv} compare \method{}-Uniform against \method{}-Anchor and \method{}-SemDiv, respectively. In \cref{fig:app_vis_anchor}, I-frame-guided selection lets \method{}-Anchor replace a weakly informative frame with the scoreboard that answers the question. In \cref{fig:app_vis_semdiv}, \method{}-SemDiv helps in a different way: it removes redundant selections and picks a set of complementary, action-centric frames, which makes the key evidence easier to find.

These examples illustrate two ways uniform Hi-I selection
can miss useful evidence: selecting an uninformative frame
or missing a brief, task-relevant event. The Lo-V stream preserves low-resolution coverage at all sampled timestamps, but recovering fine details still depends on selecting informative Hi-I frames.

{\setlength{\abovecaptionskip}{4pt}
\setlength{\belowcaptionskip}{0pt}
\begin{center}
  \begin{minipage}{0.70\linewidth}
    \centering
    \includegraphics[width=\linewidth]{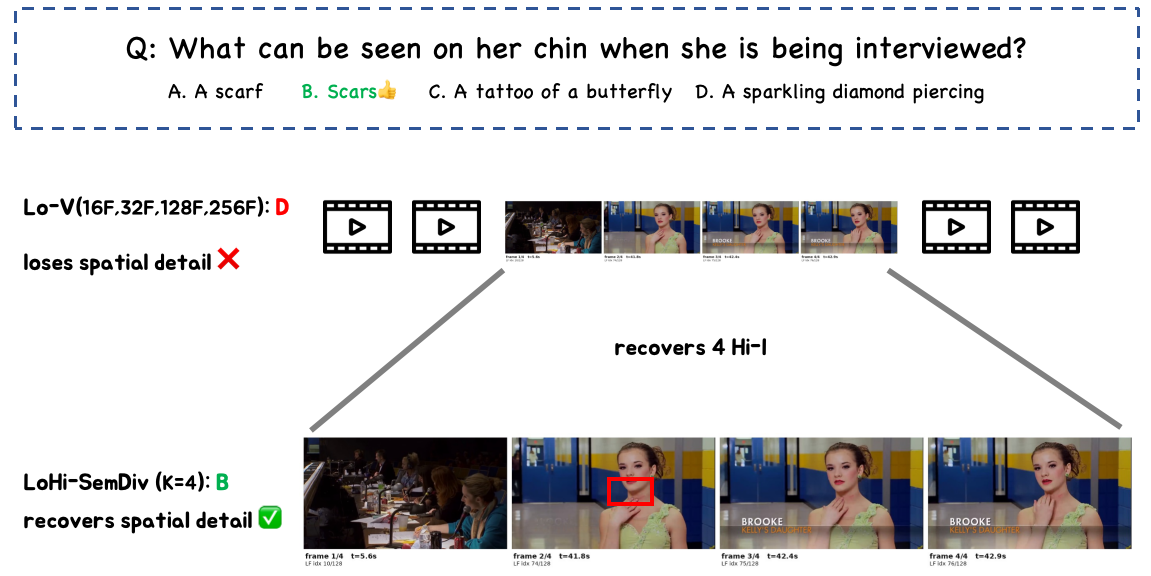}
    \captionof{figure}{Recovered Hi-I frames restore the fine-grained detail needed for the question.}
    \label{fig:app_vis_lov}
  \end{minipage}

  \par\vspace{0.75em}

  \begin{minipage}{0.70\linewidth}
    \centering
    \includegraphics[width=\linewidth]{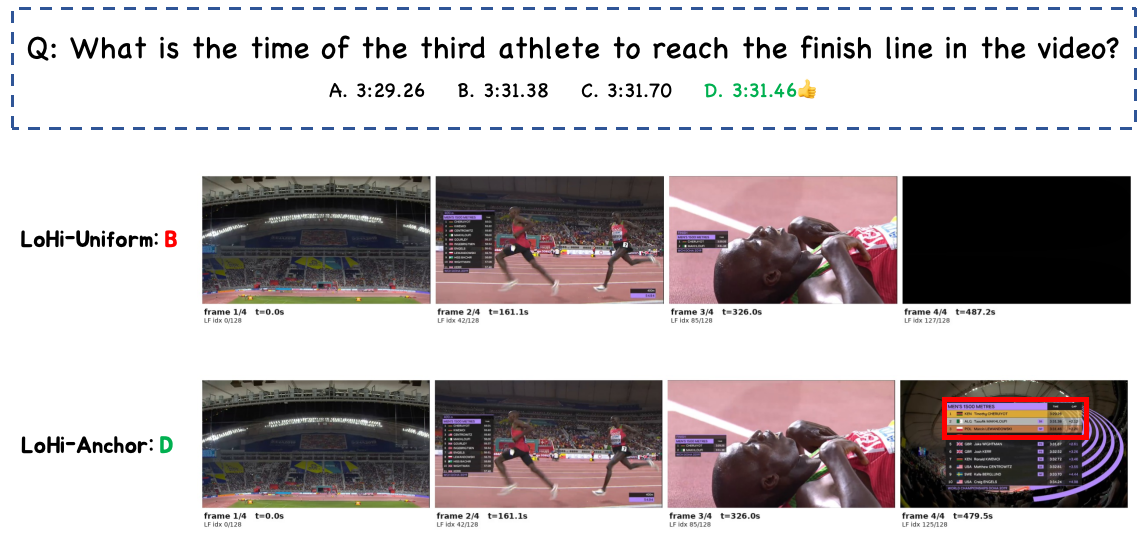}
    \captionof{figure}{LoHi-Anchor selects more informative Hi-I frames than uniform sampling.}
    \label{fig:app_vis_anchor}
  \end{minipage}

  \par\vspace{0.75em}

  \begin{minipage}{0.70\linewidth}
    \centering
    \includegraphics[width=\linewidth]{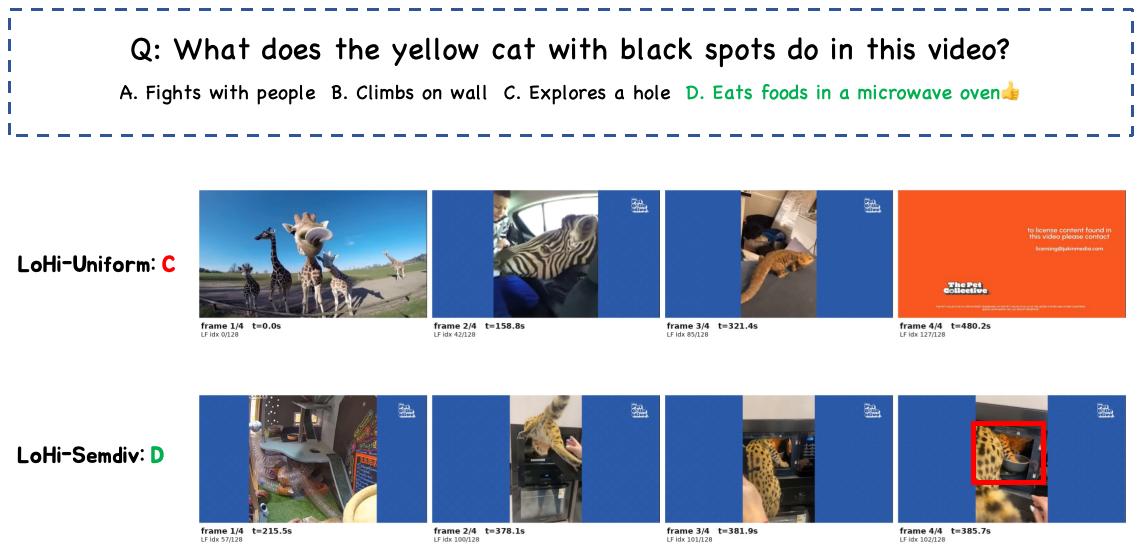}

    \vspace{-0.5em}

    \includegraphics[width=\linewidth]{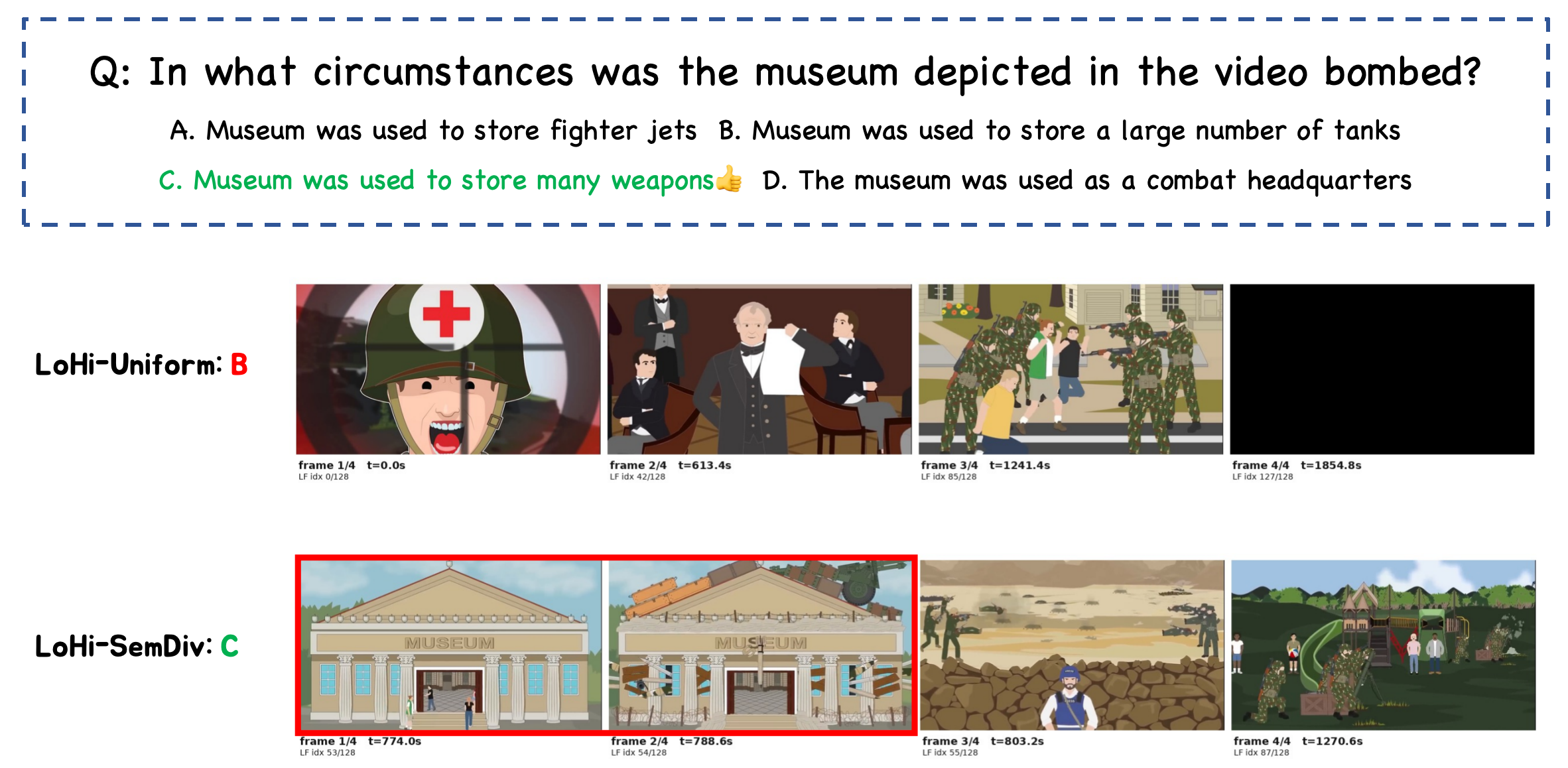}
    \captionof{figure}{LoHi-SemDiv improves Hi-I selection by favoring semantically complementary action frames.}
    \label{fig:app_vis_semdiv}
  \end{minipage}
\end{center}
}

\end{document}